\documentclass[10pt,twocolumn,letterpaper]{article}

\usepackage[pagenumbers]{cvpr} % To force page numbers, e.g. for an arXiv version

\usepackage{boldline}
\definecolor{cvprblue}{rgb}{0.21,0.49,0.74}
\usepackage[pagebackref,breaklinks,colorlinks,allcolors=cvprblue]{hyperref}
\usepackage{multirow}
\usepackage[accsupp]{axessibility}  % Improves PDF readability for those with disabilities.
\usepackage[normalem]{ulem}
\useunder{\uline}{\ul}{}

\def\paperID{13643} % *** Enter the Paper ID here
\def\confName{CVPR}
\def\confYear{2025}

\makeatletter
\def\blfootnote{\gdef\@thefnmark{}\@footnotetext}
\makeatother

\title{Mask$^2$DiT: Dual Mask-based Diffusion Transformer for Multi-Scene Long Video Generation}

\author{
Tianhao Qi\textsuperscript{\rm 1,2} \quad Jianlong Yuan\textsuperscript{\rm 2$^\dagger$} \quad Wanquan Feng\textsuperscript{\rm 2} \quad Shancheng Fang\textsuperscript{\rm 3*} \quad Jiawei Liu\textsuperscript{\rm 2} \quad SiYu Zhou\textsuperscript{\rm 2} \\
Qian He\textsuperscript{\rm 2} \quad Hongtao Xie\textsuperscript{\rm 1} \quad Yongdong Zhang\textsuperscript{\rm 1}\\
\textsuperscript{\rm 1}University of Science and Technology of China \quad \textsuperscript{\rm 2}Bytedance Intelligent Creation\\
\textsuperscript{\rm 3}Yuanshi Inc.\\
{\tt\small qth@mail.ustc.edu.cn \quad \{fangsc, htxie, zyd73\}@ustc.edu.cn}\\
{\tt\small \{yuanjianlong, fengwanquan, liujiawei.cc22, zhousiyu.vladimir, heqian\}@bytedance.com}
}

\begin{document}
\maketitle
\begin{abstract}
Sora has unveiled the immense potential of the Diffusion Transformer (DiT) architecture in single-scene video generation.
However, the more challenging task of multi-scene video generation, which offers broader applications, remains relatively underexplored.
To bridge this gap, we propose Mask$^2$DiT, a novel approach that establishes fine-grained, one-to-one alignment between video segments and their corresponding text annotations.
Specifically, we introduce a symmetric binary mask at each attention layer within the DiT architecture, ensuring that each text annotation applies exclusively to its respective video segment while preserving temporal coherence across visual tokens.
This attention mechanism enables precise segment-level textual-to-visual alignment, allowing the DiT architecture to effectively handle video generation tasks with a fixed number of scenes.
To further equip the DiT architecture with the ability to generate additional scenes based on existing ones, we incorporate a segment-level conditional mask, which conditions each newly generated segment on the preceding video segments, thereby enabling auto-regressive scene extension.
Both qualitative and quantitative experiments confirm that Mask$^2$DiT excels in maintaining visual consistency across segments while ensuring semantic alignment between each segment and its corresponding text description.
Our project page is~\href{https://tianhao-qi.github.io/Mask2DiTProject/}{https://tianhao-qi.github.io/Mask2DiTProject/}.
\end{abstract}
\blfootnote{$\dagger$Project lead *Corresponding author.}    
\vspace{-5mm}
\section{Introduction}
\label{sec:intro}
Recently, the release of Sora~\cite{brooks2024video} has sparked significant enthusiasm within the research community for text-to-video (T2V) models based on the scalable DiT~\cite{peebles2023scalable} architecture, challenging the dominance of the earlier U-Net~\cite{ronneberger2015u} framework.
Advances in T2V models have demonstrated remarkable success in generating single-scene videos, highlighting their potential for producing visually compelling and engaging content.
However, from a practical standpoint, the ability to generate coherent multi-scene videos is crucial for broader applications, including film production, educational content creation, and immersive virtual experiences.

To this end, early research on multi-scene long video generation based on U-Net architecture~\cite{ramos2024contrastive, Long:ECCV24, lin2023videodirectorgpt, yuan2024mora, zhou2024storydiffusion, oh2025mevg, bansal2024talc} has consistently adopted multiple prompts.
In these approaches, distinct prompts are used to generate different scenes within a video, with each scene's visual content conditioned on a specific prompt that describes it individually, rather than relying on a single prompt to represent all scenes collectively.
A representative line of work~\cite{Long:ECCV24, lin2023videodirectorgpt,yuan2024mora} follows this paradigm by applying a pre-trained single-scene T2V model to generate separate video segments for each prompt, which are then directly concatenated to form a multi-scene video.
However, this strategy often results in visual discontinuities and abrupt scene transitions due to the independent generation of each segment without considering global temporal coherence.
To address this, subsequent research explores training-free~\cite{oh2025mevg} and fine-tuning-based~\cite{bansal2024talc,ramos2024contrastive} techniques that explicitly account for inter-segment temporal coherence.
While such methods improve continuity, they typically limit the number of frames per scene, as the U-Net architecture lacks scalability in modeling long-range temporal dependencies, thereby constraining each segment’s ability to fully convey its narrative content.
Alternatively, another approach~\cite{zhou2024storydiffusion} employs text-to-image (T2I) models enhanced with a consistent self-attention mechanism to produce a series of coherent keyframes, followed by an image-to-video (I2V) model that sequentially connects these keyframes to construct a multi-scene video.
However, we observe that the keyframe synthesis process fails to sufficiently consider the temporal positioning of each keyframe within the overall video timeline and the associated motion dynamics.
As a result, connecting these keyframes through the I2V model often leads to unnatural motion transitions.
Furthermore, the synthesized keyframes may lack sufficient spatial information, resulting in visual inconsistencies across scenes.
In summary, current approaches to multi-scene video generation primarily rely on off-the-shelf T2V and I2V models built upon the U-Net~\cite{ronneberger2015u} architecture.
In contrast, research on developing an effective DiT~\cite{peebles2023scalable} architecture tailored for multi-scene video generation remains in its early stages.

Concerning the above problems, we pioneeringly propose Mask$^2$DiT to address the challenge of multi-scene long video generation, built upon the DiT architecture and grounded in the core principle of \textbf{multi-prompt guidance}.
Specifically, we begin by focusing on generating videos with a fixed number of scenes, ensuring that the duration of each scene is consistent with that supported by the pre-trained T2V model.
This is implemented by concatenating the textual and visual tokens from multiple scenes into a unified one-dimensional sequence.
However, such direct concatenation disrupts the fine-grained alignment between individual scenes and their corresponding text annotations.
To resolve this issue, we introduce a symmetric binary mask at each attention layer within the DiT architecture.
This mask enforces that each text annotation attends only to its corresponding video segment, thereby restoring the one-to-one alignment between video segments and their associated textual descriptions.
Simultaneously, to preserve visual coherence, we maintain attention modeling within all video segments, enabling consistent interactions across visual tokens.
With the incorporation of this attention mask, we effectively adapt the pre-trained DiT-based T2V model for video generation tasks involving a fixed number of scenes.

Furthermore, we equip the model with an auto-regressive scene extension capability, enabling it to generate additional scenes beyond the initial fixed number.
Specifically, we introduce a segment-level conditional mask to perform denoising training on the final video segment, using all preceding segments as contextual conditions.
During training, this conditional scene prediction task is interleaved with the standard multi-scene video generation task, where no conditional mask is applied, at a predefined ratio.
To reduce reliance on a large number of consecutive video segments with strong contextual continuity, we additionally design a pre-training task using non-contiguous video segments concatenated into longer videos.
In this setting, the conditional mask is omitted due to the lack of inter-segment coherence.
To the best of our knowledge, this work is the first to integrate the core principle of \textbf{multi-prompt guidance} within the DiT architecture.
Extensive experiments validate the effectiveness of Mask$^2$DiT in generating multi-scene videos with strong visual consistency across segments and accurate semantic alignment within each segment.

Overall, our contributions are threefold:
\begin{itemize}
    \item We introduce a symmetric binary mask at each attention layer within the DiT architecture, enabling the generation of a fixed number of visually consistent scenes.
    \item We incorporate a segment-level conditional mask for scene prediction training, equipping the T2V model with auto-regressive scene extension capabilities to synthesize additional scenes beyond the initial fixed number.
    \item We design a pre-training task using non-contiguous video segments to reduce reliance on large-scale consecutive video data with strong contextual coherence.
\end{itemize}
\vspace{-3mm}
\section{Related Work}
\vspace{-1mm}
\label{sec:related}
\subsection{Diffusion-based Video Generation}
T2V generation poses challenges in achieving both visual realism and semantic alignment with text prompts.
Early methods such as Imagen Video~\cite{ho2022imagen}, Make-A-Video~\cite{singer2022make}, and Lumiere~\cite{bar2024lumiere} produce low-resolution videos, which are subsequently enhanced by spatial and temporal super-resolution.
However, these multi-stage pipelines often incur substantial computational overhead and require extensive hyperparameter tuning.
To improve efficiency, video auto-encoders are introduced.
Methods like VideoLDM~\cite{blattmann2023align} and Phenaki~\cite{villegas2022phenaki} map videos into compact latent spaces, while LVDM~\cite{he2022latent} further optimizes computational efficiency.
More recent approaches adapt T2I models for T2V generation, as demonstrated by AnimateDiff~\cite{guo2023animatediff} and VideoCrafter~\cite{chen2023videocrafter1, chen2024videocrafter2}, though these models remain limited by the U-Net architecture.
In contrast, transformer-based models such as DiT~\cite{peebles2023scalable}, Sora~\cite{brooks2024video}, and VideoPoet~\cite{kondratyuk2023videopoet} overcome U-Net’s limitations, offering greater scalability and improved generation quality. MovieGen~\cite{polyak2024movie} extends this direction by generating high-quality 1080p videos with content control, while open-source projects like Open-Sora\cite{opensora} and Open-Sora-Plan~\cite{pku_yuan_lab_and_tuzhan_ai_etc_2024_10948109} contribute to accessibility and reproducibility.
CogVideoX~\cite{yang2024cogvideox} adopts a multi-modal DiT architecture in which text prompts and video content are encoded separately and then fused by a stack of expert transformer blocks, ensuring precise alignment between textual input and generated video content.
While these methods excel in single-scene generation, they remain limited in handling multi-scene video synthesis.
To this end, we extend CogVideoX~\cite{yang2024cogvideox} to support multi-scene video generation.
In parallel, recent advances in disentangled representation learning~\cite{qi2024deadiff}, dynamic transformer architectures~\cite{lu2025dhvt}, and iterative language modeling~\cite{fang2022abinet++} have shown promise across various visual and multimodal tasks.
These developments inspire our design choices in building a scalable, transformer-based framework for coherent and semantically aligned video generation.

I2V generation uses static images as anchors, with accompanying text guiding the motion dynamics.
VideoCrafter~\cite{chen2023videocrafter1, chen2024videocrafter2} and SVD~\cite{blattmann2023stable} introduce CLIP embeddings to enhance cross-frame consistency, while Animate Anyone~\cite{hu2024animate} further improves detail preservation.
Subsequent advancements, including SEINE~\cite{chen2023seine} and PixelDance~\cite{zeng2024make}, expand input channels to better retain visual fidelity.
Transformer-based methods~\cite{opensora, yang2024cogvideox, polyak2024movie, pku_yuan_lab_and_tuzhan_ai_etc_2024_10948109} have also advanced the quality and coherence.
However, the aforementioned I2V models exhibit limited capability in mitigating visual inconsistencies across the initial frames of individual scenes in multi-scene videos.

\subsection{Multi-Scene Video Generation}
Numerous methods~\cite{ramos2024contrastive, Long:ECCV24, lin2023videodirectorgpt, yuan2024mora, zhou2024storydiffusion, oh2025mevg, bansal2024talc} have been proposed to generate coherent multi-scene videos.
VideoStudio~\cite{Long:ECCV24} and VideoDirectorGPT~\cite{lin2023videodirectorgpt} leverage large language models (LLMs) to create multi-scene video scripts, serving as structured guides for downstream video generation processes.
Similarly, Mora~\cite{yuan2024mora} adopts a multi-agent strategy to divide the generation process into scene-level tasks.
However, these approaches struggle to ensure global visual coherence across scenes.
To address this, StoryDiffusion~\cite{zhou2024storydiffusion} incorporates consistent self-attention and smooth transition modules to enhance frame continuity. MEVG~\cite{oh2025mevg} iteratively refines latent vectors to maintain visual and dynamic consistency throughout the video, while TALC~\cite{bansal2024talc} aligns video scenes with sequential text descriptions, ensuring visual coherence in multi-scene outputs.
In contrast to TALC~\cite{bansal2024talc}, our approach leverages a transformer-based architecture with distinct prompts and attention mechanisms across segments, enabling coherent multi-scene video generation with fine-grained textual alignment and support for auto-regressive scene extension.
\begin{figure*}[t]
    \centering
    \includegraphics[width=1.0\linewidth]{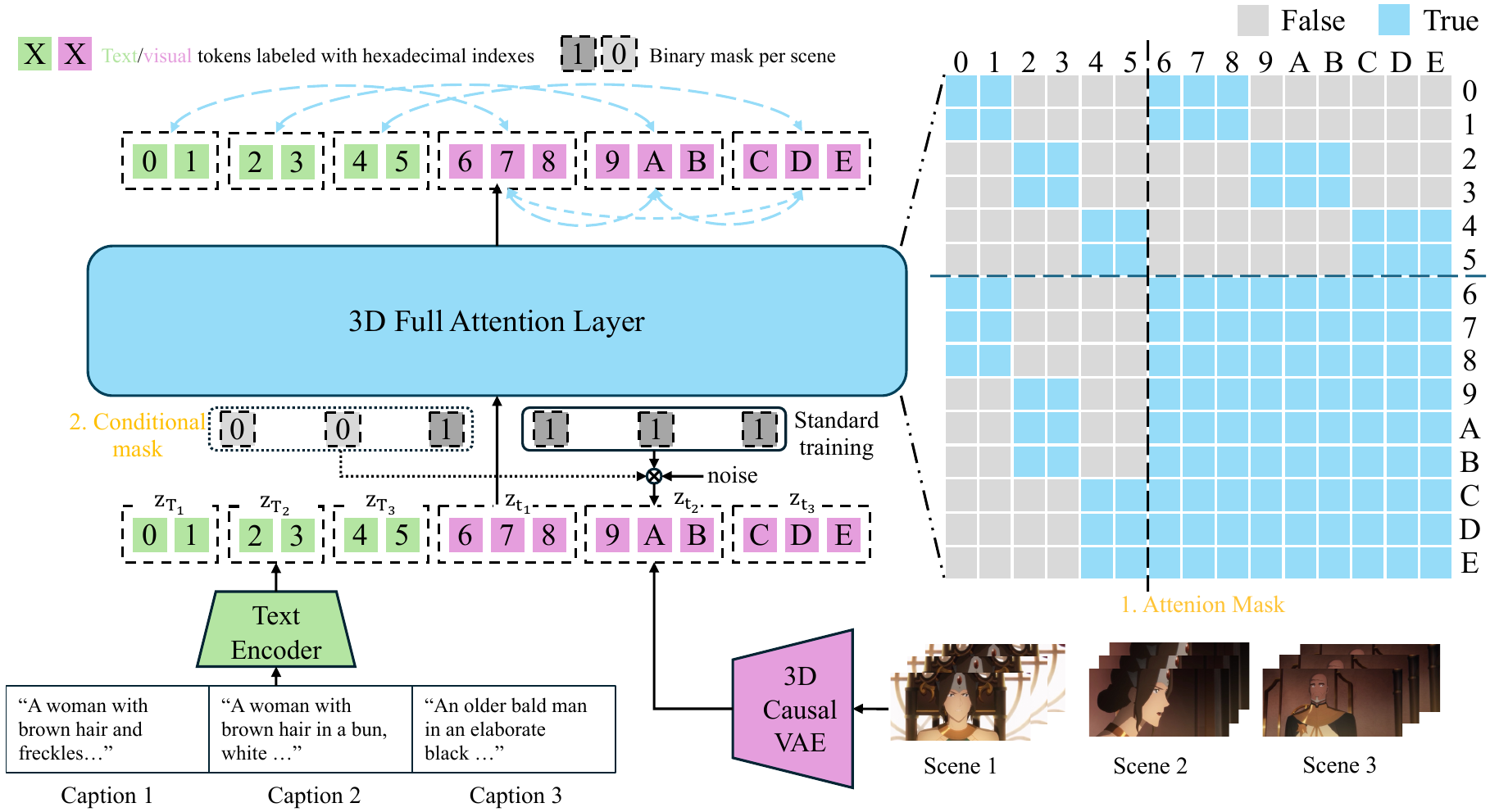}
    \vspace{-7mm}
    \caption{The overall pipeline of our method Mask$^2$DiT. First, we concatenate the text and video token sequences of $n$ scenes in temporal order, where $n=3$ is a fixed constant. The text tokens (indices 0-5) are placed at the beginning, followed by the video tokens (indices 6-E). Then, we introduce a symmetric binary attention mask to ensure that text annotations affect only its corresponding video frame range, while still preserving temporal continuity across all visual tokens. Finally, we introduce a segment-level conditional mask that predicts the final segment based on the preceding $n-1$ video segments, equipping the model with the capability to extend scenes auto-regressively.}
    \vspace{-3mm}
    \label{fig:pipeline}
\end{figure*}

\vspace{-7mm}
\section{Method}
\vspace{-2mm}
\subsection{Preliminary}
\label{sec:preliminary}
\vspace{-1mm}
In light of the Diffusion Transformer's excellent scalability and impressive performance in visual content generation, we adopt the open-sourced CogVideoX~\cite{yang2024cogvideox} as the base model for implementing our approach.
In line with mainstream methodologies, CogVideoX encodes the input video $x$ into a one-dimensional visual token sequence $z_{V}$ using a 3D Causal VAE.
The denoising network $v_\theta$, implemented as an encoder-only Transformer, then receives a one-dimensional mixed token sequence consisting of text tokens $z_{T}$ from a text encoder and visual tokens $z_t=\sqrt{\bar{\alpha}_t}\epsilon+\sqrt{1-\bar{\alpha}_t}z_{V}$, which are generated by perturbing $z_{V}$ with a random Gaussian noise $\epsilon$ according to the zero SNR schedule~\cite{lin2024common} at time $t$.
During training, the model adopts the v-prediction~\cite{salimans2022progressive} objective, formulated as:
\begin{equation}
    L = \mathbb{E}_{z_{V}, z_{T}, \epsilon \sim \mathcal{N}(0,1), t}\left[\left\|v-v_{\theta}\left(z_{t}, t, z_{T}\right)\right\|_{2}^{2}\right],
    \label{equ:1}
\end{equation}
where $v=\sqrt{\bar{\alpha}_t}\epsilon-\sqrt{1-\bar{\alpha}_t}z_{V}$ represents the velocity.

\begin{figure*}
    \centering
    \includegraphics[width=1.0\linewidth]{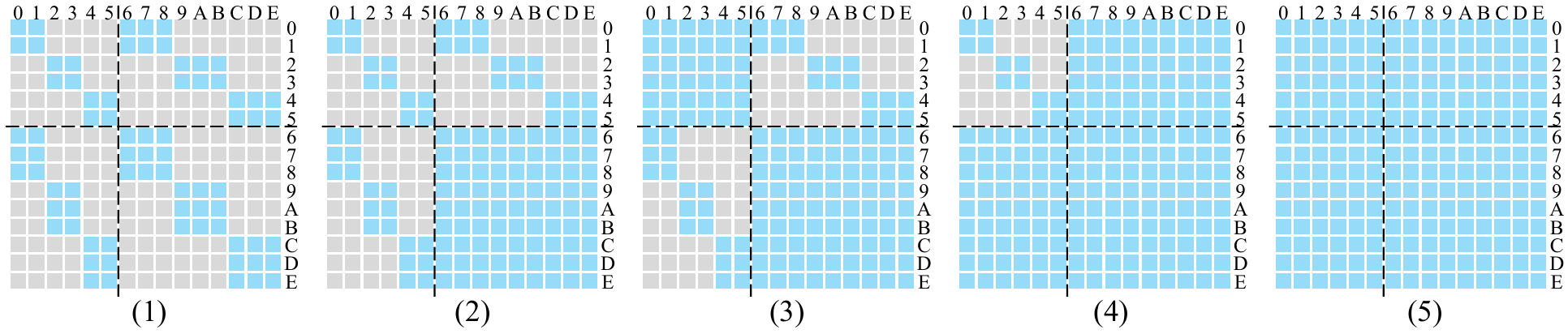}
    \vspace{-7mm}
    \caption{Attention mask variants for exploration. The first configuration serves as the baseline. The second is the most computationally efficient, achieving a balance between semantic consistency by aligning each video segment with its corresponding text and visual consistency across all segments. The three rightmost configurations are further explored to capture additional inter-token correlations.}
    \label{fig:attention_masks}
    \vspace{-5mm}
\end{figure*}

\subsection{Symmetric Binary Attention Mask}
\label{sec:attn_msk}
As discussed in~\cref{sec:intro}, we first focus on extending the pre-trained CogVideoX to generation tasks involving a fixed number of $n$ scenes, ensuring that the duration of each scene is consistent with that supported by the original model.

To this end, we follow the core principle of \textbf{multi-prompt guidance} and concatenate the text and video token sequences of the $n$ scenes in temporal order, positioning the text tokens before the video tokens, \textit{i.e.}, $[z_{T_1}, \dots, z_{T_n}; z_{t_1}, \dots, z_{t_n}]$, as shown in~\cref{fig:pipeline}.
However, straightforward token concatenation disrupts the fine-grained alignment between each scene and its corresponding text annotation.
To address this, we propose introducing an attention mask $m_{attn}$ to restore the fine-grained, one-to-one alignment.
To validate our hypothesis, we design multiple variants of attention masks, as illustrated in~\cref{fig:attention_masks}.
The first configuration is equivalent to jointly processing $n$ video-text pairs within a single batch, where attention is restricted to the corresponding text and video tokens of each pair, serving as a baseline for comparison.
To maintain visual coherence across video segments, we argue that modeling correlations solely between different visual segments is sufficient, as illustrated in~\cref{fig:attention_masks} (2).
The three rightmost attention masks in~\cref{fig:attention_masks} are further explored to capture additional inter-token correlations.
With the incorporation of attention masks, each scene segment is able to attend more effectively to its corresponding text prompt.
This hypothesis is empirically validated by our experimental results.

\begin{figure}
    \centering
    \includegraphics[width=1.0\linewidth]{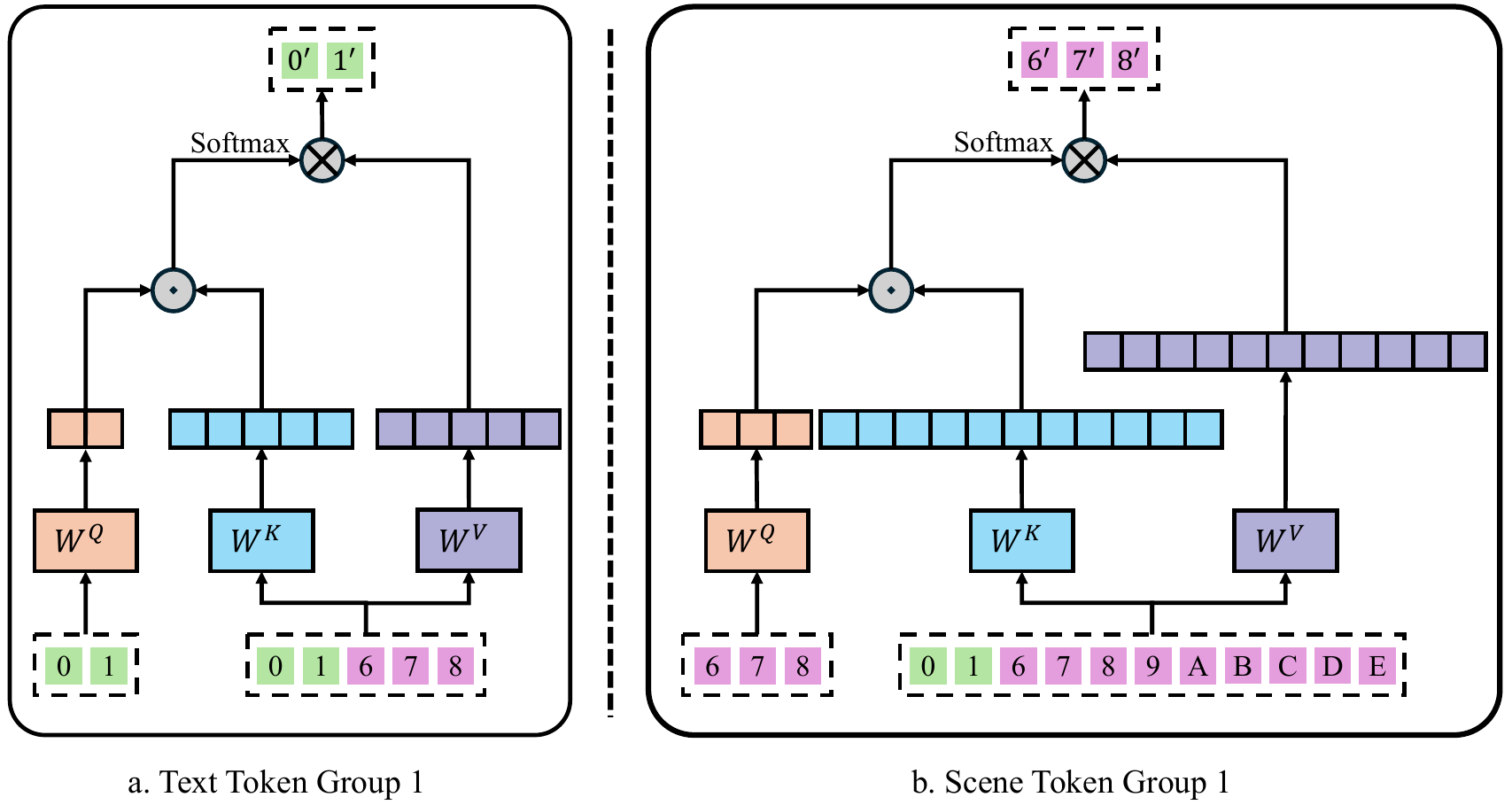}
    \vspace{-7mm}
    \caption{The illustration of our grouped attention mechanism for Text Token Group 1 and Scene Token Group 1. For Text Token Group 1, the Query consists of text tokens from the first scene segment, while the Key and Value are formed by concatenating text and video tokens of the same scene segment. For Scene Token Group 1, the Query is video tokens from the first scene segment, and the Key and Value are obtained by concatenating text tokens of the first scene segment with the video tokens from all scenes.}
    \label{fig:gam}
    \vspace{-5mm}
\end{figure}

\noindent\textbf{Grouped Attention Mechanism}: Noteworthy, the total length of the concatenated token sequence is considerable, therefore, the memory coverage of the attention mask becomes significant.
To address this, we implement a grouped attention mechanism in our practical setup.
Taking the attention mask we utilize in~\cref{fig:pipeline} for example, we divide the entire one-dimensional token sequence into $2n$ groups, consisting of $n$ scene token groups and their corresponding $n$ text token groups, as demonstrated in~\cref{fig:gam}.
We treat each group of tokens as query tokens and identify the corresponding key and value tokens in the attention mask, followed by performing a cross-attention calculation.
In this manner, a single self-attention calculation is replaced by $2n$ instances of cross-attention calculations. By eliminating the need for attention masks and reducing redundant computations between certain tokens, both memory usage and computational efficiency are significantly improved.

\vspace{-2mm}
\subsection{Segment-level Conditional Mask}
\label{sec:cmask}
As described in~\cref{sec:attn_msk}, the pre-trained T2V generation model can be adapted to tasks involving a fixed number of scenes.
Nevertheless, when the number of scenes becomes excessively large, the time and resource demands for both training and inference grow prohibitively high.
To enhance the model’s scalability, we introduce a segment-level conditional mask $m_{c}$ that enables auto-regressive scene extension, as illustrated in~\cref{fig:pipeline}.
Specifically, during training, we randomly drop the Gaussian noise applied to the visual tokens of the first $n-1$ segments, allowing them to serve as contextual input for denoising the $n$-th segment.
The loss is computed solely over the visual tokens of the $n$-th video segment.
Formally, the optimization objective is defined as:
\begin{equation}
    L = \mathbb{E}_{z_{V}, z_{T}, \epsilon \sim \mathcal{N}(0,1), t}\sum_{i=1}^{n}\left[m_{c_i}\cdot\left\|v-v_{\theta}\left(z_{t_i}, t, z_{T_i}\right)\right\|_{2}^{2}\right],
    \label{equ:2}
\end{equation}
where the first $n-1$ elements of $m_{c}$ are set to 0, and the $n$-th element is set to 1.
After training, the model is capable of extending video scenes in an auto-regressive manner.

\subsection{Training Process}
Our training pipeline consists of two main stages: pre-training and supervised fine-tuning.

During the pre-training stage, we concatenate $n$ single-scene videos without contextual relationships into a single long video, treating it as one training sample.
In this stage, we apply only the attention mask $m_{attn}$ and perform standard training without using the segment-level conditional mask, under the supervision of~\cref{equ:3},
\begin{equation}
    L = \mathbb{E}_{z_T, z_{V}, \epsilon \sim \mathcal{N}(0,1), t}\sum_{i=1}^{n}\left[\left\|v-v_{\theta}\left(z_{t_i}, t, z_{T_i}\right)\right\|_{2}^{2}\right],
    \label{equ:3}
\end{equation}
aiming to adapt the model to multi-scene generation scenarios with a frame length scaled by a factor of $n$.
By default, we adopt the second variant, as shown in~\cref{fig:attention_masks}(2).

In the supervised fine-tuning stage, we train the model on videos containing $n$ consecutive scenes with contextual relationships.
This stage is designed to enhance the model's ability to generate videos with improved visual coherence across scenes and to support auto-regressive scene extension.
To this end, we introduce the segment-level conditional mask $m_{c}$, as described in~\cref{sec:cmask}, and apply it with a probability $p$ under the supervision of~\cref{equ:2}.
With the remaining probability $1-p$, training follows the objective defined in~\cref{equ:3}.
Notably, this two-stage training strategy significantly reduces the dependency on labor-intensive video datasets composed of $n$ consecutive scenes with contextual relationships, which would otherwise be required throughout the entire training process.
\vspace{-3mm}
\section{Experiment}
\vspace{-2mm}
\subsection{Experiment Settings}
\noindent\textbf{Implementation Details.}
As outlined in~\cref{sec:preliminary}, we adopt the open-sourced CogVideoX~\cite{yang2024cogvideox} as the base model for T2V generation.
To balance computational efficiency and model scalability, we fix the number of scenes to $n = 3$.
During training, the model undergoes 10000 iterations of pre-training, followed by 10000 steps of supervised fine-tuning.
The batch size is set to 8 throughout, with each video rendered at a resolution of 480×720 and consisting of 49 frames per scene, resulting in 147 frames per training sample.
The learning rate and the probability $p$ are set to 1e-5 and 0.5, respectively.

\noindent\textbf{Pre-Training Dataset.} Given the scarcity of coherent multi-scene video data and the relative abundance of single-scene data (e.g., Panda70M~\cite{chen2024panda}), we select 1 million video samples from Panda70M~\cite{chen2024panda} to train the Mask$^2$DiT model.
To simulate multi-scene coherence during training, we randomly sample three single-scene video clips from the dataset and concatenate them to construct a composite multi-scene video.

\begin{figure*}
    \centering
    \includegraphics[width=0.75\linewidth]{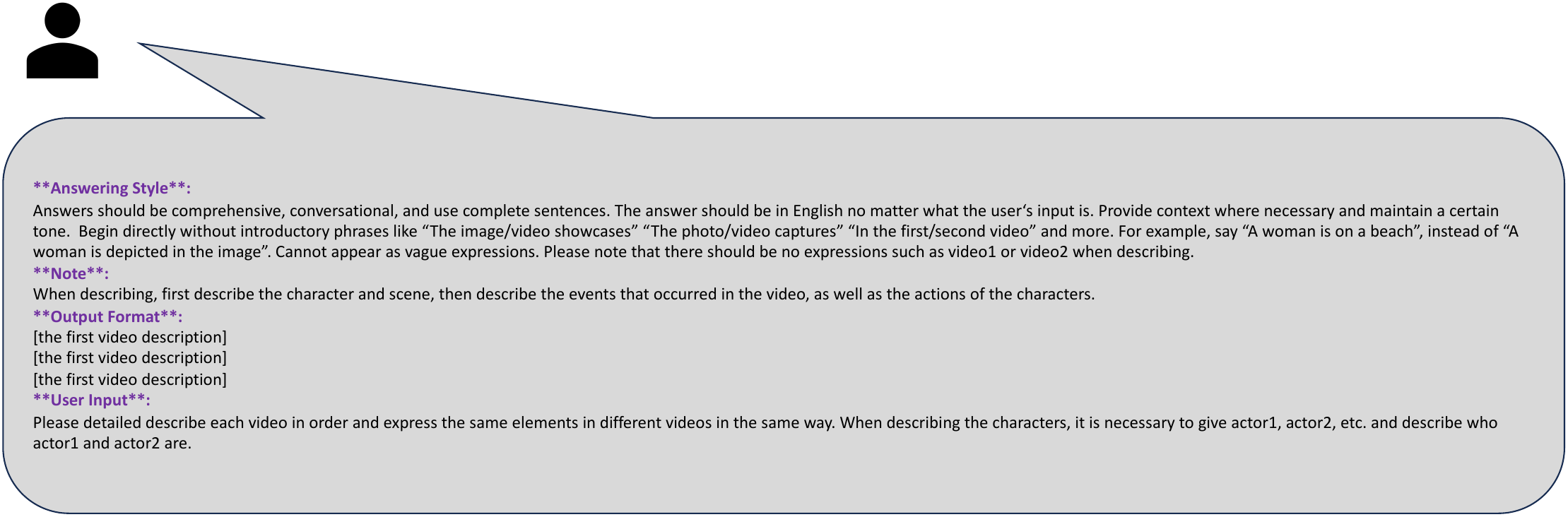}
    \vspace{-3mm}
    \caption{Illustration of the supervised fine-tuning dataset construction process. Long-form videos are segmented into 10-minute clips, then divided into individual shots using PySceneDetect~\cite{pyscenedetect}. Shot combinations are filtered based on inter-video relevance, computed using ViClip~\cite{wang2023internvid}. Descriptive annotations that preserve character and background consistency are generated using Gemini~\cite{team2023gemini}.}
    \label{fig:train_dataset}
    \vspace{-4mm}
\end{figure*}

\noindent\textbf{Supervised Fine-tuning Dataset (SFT).}
To enhance scene-to-scene coherence, we construct a training dataset composed of content-relevant video combinations.
We first collect a large corpus of long-form videos, segment them into 10-minute clips, and apply PySceneDetect~\cite{pyscenedetect} to divide each clip into individual shots.
After filtering, only combinations of shots with sufficient inter-video relevance, measured using the ViClip~\cite{wang2023internvid} model, are retained.
Combinations are selected based on a predefined similarity threshold.
For the final dataset, we generate descriptive annotations using Gemini~\cite{team2023gemini}, ensuring consistency in both character identity and background appearance, as illustrated in Fig~\ref{fig:train_dataset}.

\begin{figure*}[tb]
    \centering
    \includegraphics[width=0.75\linewidth]{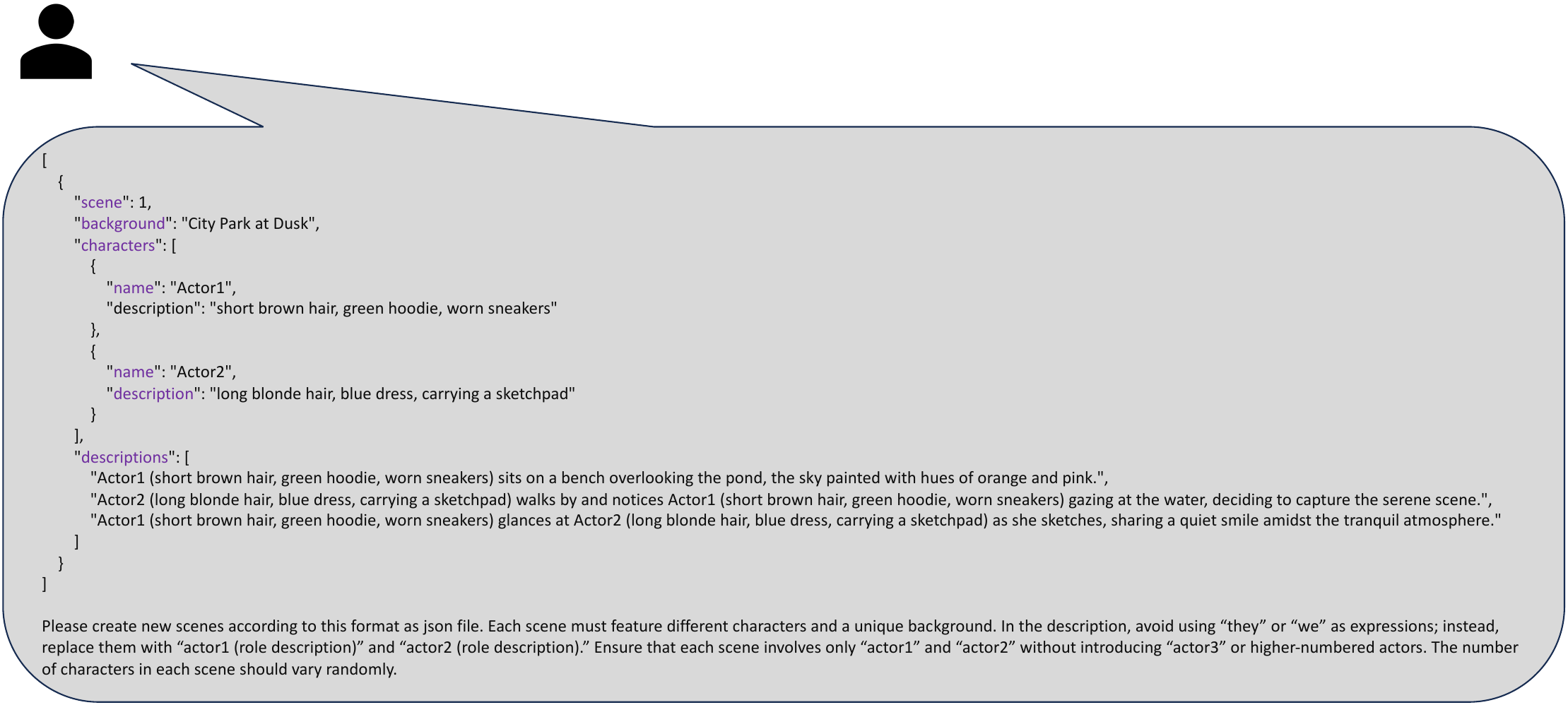}
    \vspace{-3mm}
    \caption{Structure of the evaluation dataset. It contains 50 scenes, each accompanied by three prompts featuring a variable number of characters. ChatGPT~\cite{openai2024chatgpt} was used to generate diverse prompts based on this structure, enabling comprehensive evaluation of video generation models in multi-scene scenarios.}
    \label{fig:eval_dataset}
    \vspace{-5mm}
\end{figure*}

\noindent\textbf{Evaluation Dataset.}
To evaluate the effectiveness of our approach, we construct an evaluation dataset comprising 50 scenes, each paired with three prompts containing a variable number of characters.
To ensure prompt diversity and broad scene coverage, we leverage ChatGPT~\cite{openai2024chatgpt} to generate varied scene descriptions following the structure illustrated in Fig.~\ref{fig:eval_dataset}.
This dataset, comprising 5,371 multi-scene videos, serves as a standardized test set for assessing the performance of video generation models in multi-scene scenarios.

\noindent\textbf{Evaluation Metrics. }
To assess the quality of video generation results, we evaluate three key aspects: Visual Consistency, which measures the coherence of entities and backgrounds across scenes; Semantic Consistency, which evaluates the alignment between the generated video content and the input prompt; and Sequence Consistency, computed as the mean of the former two metrics to reflect overall coherence.
For Visual Consistency, we leverage the ViClip model~\cite{wang2023internvid} to compute video-to-video similarity.
Semantic Consistency is measured by evaluating the similarity between the generated video and the corresponding textual description. 
Additionally, we adopt the Fréchet Video Distance (FVD)~\cite{ge2024content} to assess the overall quality of the generated videos in terms of realism and distributional similarity.

\vspace{-2mm}
\subsection{Comparison with State-of-the-Arts}

\noindent\textbf{Quantitative Comparisons.}
To assess the effectiveness of our method, we compare it against several state-of-the-art (SOTA) approaches.
For CogVideoX~\cite{yang2024cogvideox}, we leverage its single-scene video generation capability by generating multiple shots using distinct prompts and concatenating them to form a multi-scene video.
For StoryDiffusion~\cite{zhou2024storydiffusion}, we utilize its keyframe generation module to produce initial frames for each scene, followed by CogVideoX’s I2V module to synthesize the remaining frames.
For TALC~\cite{bansal2024talc}, we adopt its original implementation for multi-scene video generation, which is based on a U-Net architecture.
For VideoStudio~\cite{Long:ECCV24}, we skip the script planning stage based on LLMs and instead directly use our test set to generate reference images for each scene, which are then used to render the final multi-scene video.
In contrast, our method performs direct inference over multiple scenes within the DiT architecture.

\begin{figure*}
    \centering
    \begin{subfigure}[b]{\linewidth}
        \centering
        \includegraphics[width=0.78\linewidth]{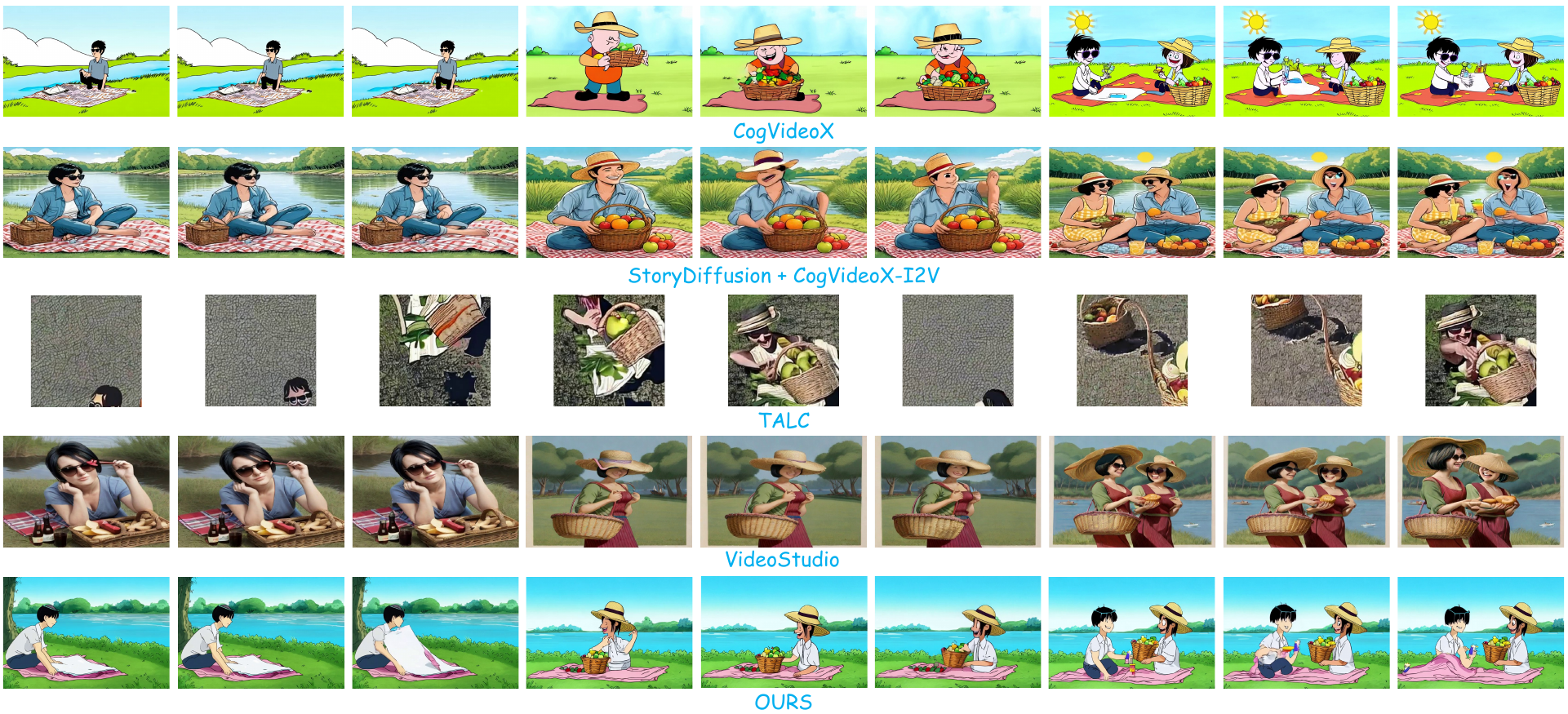}
        \caption{"Actor1 (short black hair, wearing sunglasses, spreading out a picnic blanket) sets up by the riverside, watching the gentle flow of the water.",
            "Actor2 (wearing a straw hat, carrying a basket of fruit) arrives with a cheerful smile, placing the basket on the blanket.",
            "Actor1 (short black hair, wearing sunglasses, spreading out a picnic blanket) and Actor2 (wearing a straw hat, carrying a basket of fruit) share snacks, laughing as the sun shines brightly on the river."}
        \label{fig:subfig1}
    \end{subfigure}
    % \hfill

    \begin{subfigure}[b]{\linewidth}
        \centering
        \includegraphics[width=0.78\linewidth]{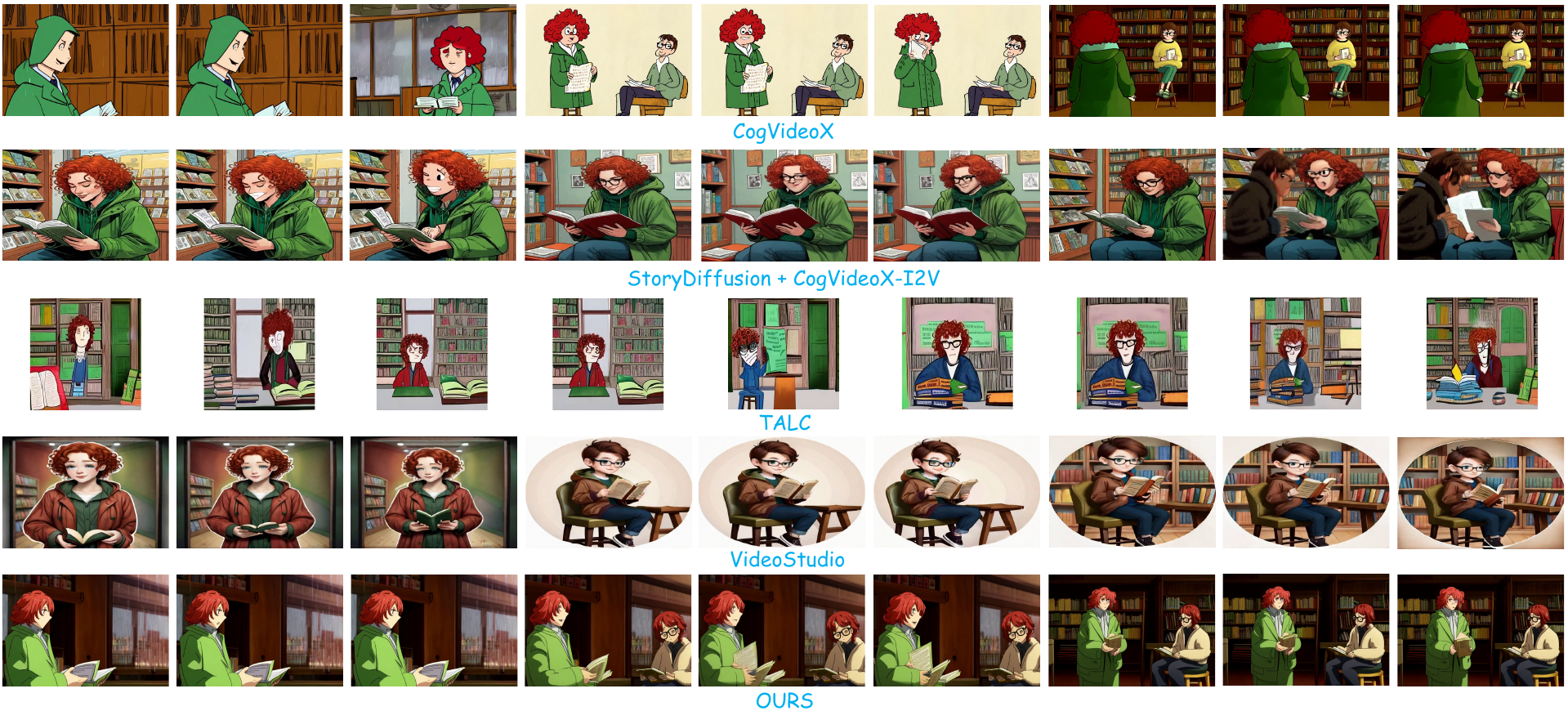}
        \caption{"Actor1 (curly red hair, green raincoat, looking at a poetry book) flips through pages with a small smile, the sound of rain tapping against the store's windows.",
            "Actor2 (wearing glasses, oversized sweater, sitting on a stool reading) notices Actor1 (curly red hair, green raincoat, looking at a poetry book) and glances up with a gentle smile.",
            "Actor1 (curly red hair, green raincoat, looking at a poetry book) approaches Actor2 (wearing glasses, oversized sweater, sitting on a stool reading), asking a question about the book in her hands, leading to a quiet conversation amid the cozy shelves."}
        \label{fig:subfig2}
    \end{subfigure}

    \begin{subfigure}[b]{\linewidth}
        \centering
        \includegraphics[width=0.78\linewidth]{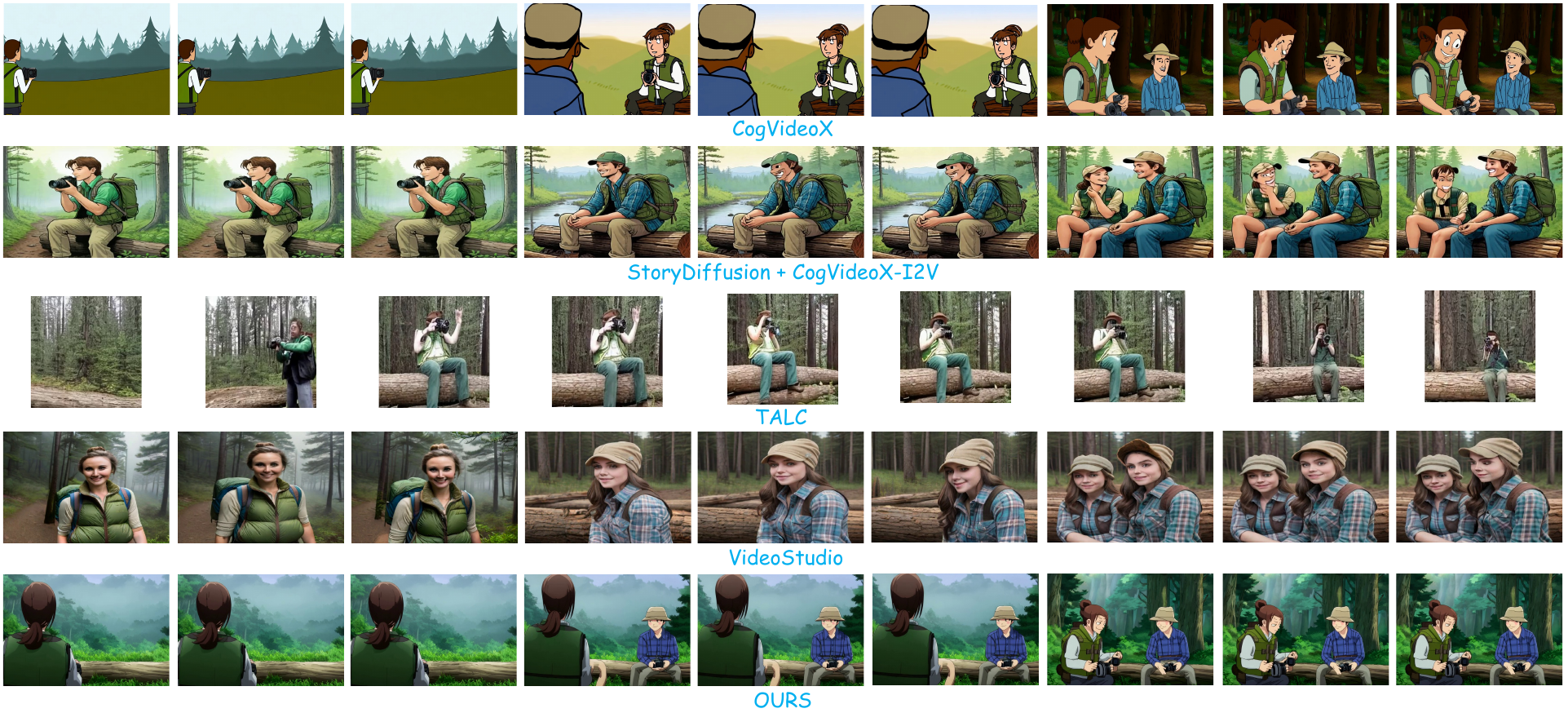}
        \caption{"Actor1 (brown hair tied back, wearing a green hiking vest, holding a camera) stands at the edge of the clearing, framing a shot of the misty forest.",
            "Actor2 (wearing a beige hat, blue flannel shirt, sitting on a log) watches Actor1 (brown hair tied back, wearing a green hiking vest, holding a camera) with a smile, appreciating the tranquility of the early morning.",
            "Actor1 (brown hair tied back, wearing a green hiking vest, holding a camera) lowers the camera and shares a quiet laugh with Actor2 (wearing a beige hat, blue flannel shirt, sitting on a log) as they both take in the serene beauty of the forest."}
        \label{fig:subfig3}
    \end{subfigure}
    \vspace{-8mm}
    \caption{Qualitative comparison with the SOTA methods. Zoom in for better visualization.}
    \label{fig:sota_show}
\end{figure*}

The experimental results are presented in Tab~\ref{tab:sota}.
Our method achieves an 8.63\% improvement over CogVideoX~\cite{yang2024cogvideox} in Sequence Consistency, particularly notable gain of 15.94\% in Visual Consistency.
This improvement is primarily attributed to our multi-scene training strategy, where attention computations explicitly model inter-scene relationships.
Likewise, our method consistently exhibits superior visual coherence in contrast with LLM-based VideoStudio~\cite{Long:ECCV24}.
Compared to the typical keyframe-based method StoryDiffusion~\cite{zhou2024storydiffusion}, which employs a T2I model followed by an I2V process, our approach yields a 1.89\% improvement in Visual Consistency.
This enhancement stems from the fact that keyframe sampling in StoryDiffusion~\cite{zhou2024storydiffusion} does not account for temporal alignment or motion continuity across scenes in multi-scene video generation.
Consequently, the subsequent I2V process often produces unnatural motion transitions, failing to preserve inter-scene coherence.
In contrast, our method establishes visual correlations directly at the model level, leading to more consistent multi-scene video generation.

Regarding TALC~\cite{bansal2024talc}, our method demonstrates a 3.48\% and 3.59\% improvement in Visual and Sequence Consistency, respectively, further validating the superiority of DiT-based architectures for multi-scene video generation over traditional U-Net-based approaches.
Moreover, our method attains the highest visual quality among all evaluated models, as evidenced by the lowest FVD score, while ranking second in Semantic Consistency, highlighting its strong alignment between visual content and prompts.
The slight gap in Semantic Consistency between our method and StoryDiffusion can be attributed to the fact that advancements in semantic consistency for text-to-video models have not yet caught up with those in text-to-image models.

\begin{table}[tbp]
    \begin{center}
    \resizebox{\linewidth}{!}{
        \begin{tabular}{l c c c c}
        \hlineB{2}
        Method  & Visual Con. (\%) & Semantic Con. (\%) & Sequence Con. (\%) & FVD ($\downarrow$) \\
        \hline\hline
        CogVideoX~\cite{yang2024cogvideox} & 55.01 & 22.64 & 38.82 & 835.35 \\
        StoryDiffusion~\cite{zhou2024storydiffusion} + CogVideoX-I2V~\cite{yang2024cogvideox} & 69.06 & \textbf{25.59} & 47.32 & 905.69 \\
        TALC~\cite{bansal2024talc} & 67.47 & 20.25 & 43.86 & 1516.59 \\
        VideoStudio~\cite{Long:ECCV24} & 61.28 & 22.64 & 41.96 & 1213.88 \\
        Ours & \textbf{70.95} & 23.94 & \textbf{47.45} & \textbf{720.01} \\
        \hlineB{2}
        \end{tabular}
    }
    \end{center}
    \vspace{-7mm}
    \caption{Quantitative comparison with the SOTA methods.}
    \label{tab:sota}
    \vspace{-7mm}
\end{table}

\begin{figure}
    \centering
    \includegraphics[width=1.0\linewidth]{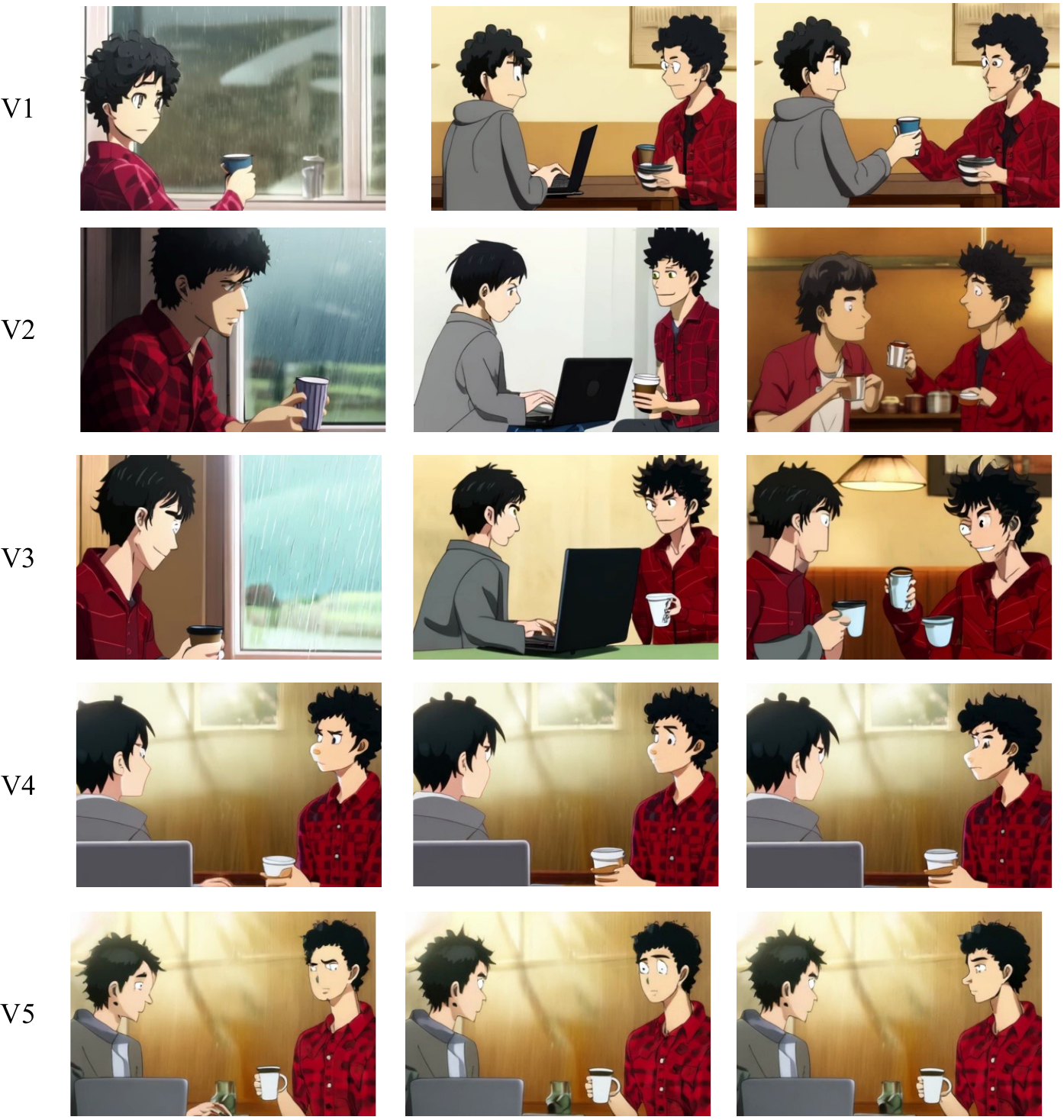}
    \vspace{-7mm}
    \caption{Qualitative comparison of attention mask mechanisms with a training batch size of 1 at a resolution of 256×384.}
    \label{fig:show_ablation_mask}
    \vspace{-3mm}
\end{figure}

\noindent\textbf{Qualitative Comparisons.}
As shown in Fig~\ref{fig:sota_show}, we conduct a qualitative comparison against the SOTA methods. The results demonstrate that our method substantially outperforms existing approaches in terms of character consistency, background coherence, and stylistic uniformity.
For example, in our generated videos, the ``boy with a straw hat" maintains consistent visual features across different shots, whereas CogVideoX and VideoStudio fail to do so due to their lack of inter-shot relational modeling, resulting in noticeable variation in visual content.
Compared to StoryDiffusion, although it retains some visual elements across shots, its keyframe-based pipeline overlooks temporal continuity and motion consistency, leading to inferior visual coherence across scenes. 
Furthermore, our method shows clear advantages over TALC, benefiting from the strong modeling capacity of the DiT architecture.

\vspace{-2mm}
\subsection{Ablation Study}
\vspace{-1mm}
To reduce computational costs, all ablation experiments are conducted at a resolution of 256×384.

\begin{table}[tbp]
\begin{center}
\resizebox{\linewidth}{!}{
\begin{tabular}{l c c c}
\hlineB{2}
Method  & Visual Con. (\%) & Semantic Con. (\%) & Sequence Con. (\%) \\
\hline\hline
v1 & 58.09 & 24.69 & 41.39 \\
v2 & 68.45 & 23.82 & 46.14 \\
v3 & 77.56 & 23.47 & 50.52 \\
v4 & 93.43 & 20.73 & 57.08 \\
v5 & 90.72 & 20.60 & 55.66 \\
\hlineB{2}
\end{tabular}
}
\end{center}
\vspace{-7mm}
\caption{Quantitative comparison of attention mask mechanisms with a training batch size of 1 at a resolution of 256×384.}
\label{tab:ablation_attn_msk}
\vspace{-5mm}
\end{table}

\noindent\textbf{Comparison of Different Attention Mask Mechanisms.}
As shown in Tab~\ref{tab:ablation_attn_msk}, we conduct comparative experiments to evaluate the impact of different variants of the attention mask.
The V2 configuration achieves a 10.36\% improvement in Visual Consistency over V1, demonstrating that inter-video attention effectively enhances Visual Consistency.
Although V3 further improves Visual Consistency compared to V2, it incurs higher computational cost due to inter-text attention and introduces the risk of information leakage between segments.

As illustrated in Fig~\ref{fig:show_ablation_mask}, the V4 and V5 attention mechanisms exhibit similar issues, including increased resource complexity and a noticeable decline in Semantic Consistency.
This reduction occurs because each prompt interacts with all video segments, which introduces significant disturbances in visual-textual alignment.
While these mechanisms enhance Visual Consistency, the gain largely results from treating all video segments as a single unified sequence, rather than preserving the structural integrity of a multi-scene video.
Based on this analysis, the V2 attention mechanism is selected as the default configuration.

\noindent\textbf{Comparison of Different Data Filtering Strategies.}
As shown in Tab~\ref{tab:ablation_dataset}, we observe that stricter data filtering criteria significantly enhance Visual Consistency and Sequence Consistency. As the dataset selection threshold increases from 0.0 to 0.8, Visual Consistency improves from 63.94\% to 73.22\%, marking a 18.21\% increase over CogVideoX which does not apply visual consistency constraints.
Similarly, Sequence Consistency steadily rises, reaching 48.38\% at the highest filtering threshold, 9.56\% higher than CogVideoX.
These findings suggest that refining dataset selection strengthens the model’s ability to maintain visual coherence and temporal consistency across scenes.
However, a trade-off is observed in Semantic Consistency, where the highest score (24.31\%) is achieved with the least restrictive filtering (Dataset-0.0), while stricter criteria lead to a slight decline. This suggests that although improved dataset quality enhances visual and sequential coherence, it may reduce textual fidelity by eliminating semantically rich but diverse samples.

\begin{table}[tbp]
\begin{center}
\resizebox{\linewidth}{!}{
\begin{tabular}{l c c c}
\hlineB{2}
Method  & Visual Con. (\%) & Semantic Con. (\%) & Sequence Con. (\%)\\
\hline\hline
CogVideoX~\cite{yang2024cogvideox}  & 55.01 & 22.64 & 38.82 \\
Ours + Dataset-0.0 & 63.94 & \textbf{24.31} & 44.13 \\
Ours + Dataset-0.6 & 66.46 & 23.96 & 45.21  \\
Ours + Dataset-0.7 & 67.71 & 23.98 & 45.85 \\
Ours + Dataset-0.8 & \textbf{73.22} & 23.54 & \textbf{48.38} \\
\hlineB{2}
\end{tabular}
}
\end{center}
\vspace{-7mm}
\caption{Quantitative comparison of data filtering strategies with a training batch size of 4 at a resolution of 256×384.}
\label{tab:ablation_dataset}
\vspace{-7mm}
\end{table}

\vspace{-3mm}
\section{Conclusion}
\vspace{-2mm}

We present Mask$^2$DiT, a novel dual-mask-based diffusion transformer designed to advance multi-scene long video generation.
By introducing a symmetric binary mask for fine-grained alignment between text annotations and corresponding video segments, along with a segment-level conditional mask for auto-regressive scene extension, Mask$^2$DiT enhances temporal coherence and semantic consistency in generated videos.
Our approach achieves notable gains in visual and semantic consistency compared to the SOTA methods, enabling high-quality, coherent multi-scene videos without sacrificing scene-specific details.
Future directions include enhancing the model's generalization to real-world video data, supporting adaptive-length scenes, and scaling up the total video duration and resolution.\looseness=-1

\section*{Acknowledgement}
This work is supported by the National Nature Science Foundation of China (62425114, 62121002, U23B2028, 62232006).
We acknowledge the support of GPU cluster built by MCC Lab of Information Science and Technology Institution, USTC.
We also thank the USTC supercomputing center for
providing computational resources for this project.
Special thanks to Gen Li~\footnote{ligen.lab@bytedance.com}, Yaqi Cai~\footnote{cyaqi@mail.ustc.edu.cn} and Yanmei Tian~\footnote{tianym123@gmail.com} for their valuable contributions to this research.
{
    \small
    \bibliographystyle{ieeenat_fullname}
    \bibliography{main}
}

\clearpage

\clearpage
\setcounter{page}{1}
\maketitlesupplementary
\section{Supplementary}

\begin{figure*}[htpb]
    \centering
    \begin{subfigure}[b]{\linewidth}
        \centering
        \includegraphics[width=\linewidth]{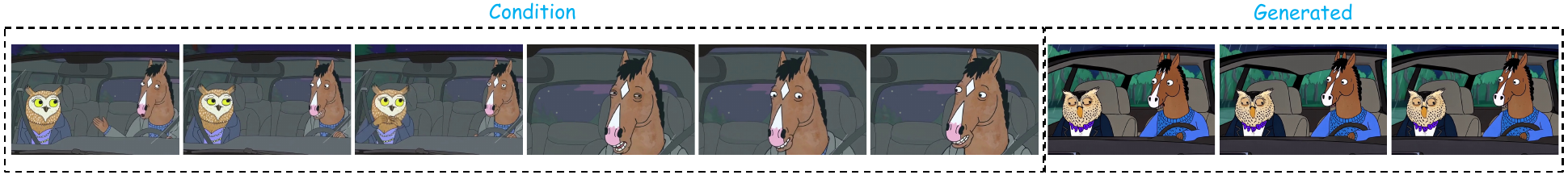}
        \caption{
        "An owl and a horse are driving in a car at night. The owl gestures while talking to the horse, who is focused on driving.", "A cartoon horse in a gray sweater sits in a car at night, looking around as if reacting to something off-screen.", "An owl in a dark jacket and purple necklace sits anxiously in the passenger seat of a car driven by a horse in a blue sweater. They drive through a rainy, wooded area at night."
        }
        \label{fig:subfig11}
    \end{subfigure}
    % \hfill

    \begin{subfigure}[b]{\linewidth}
        \centering
        \includegraphics[width=\linewidth]{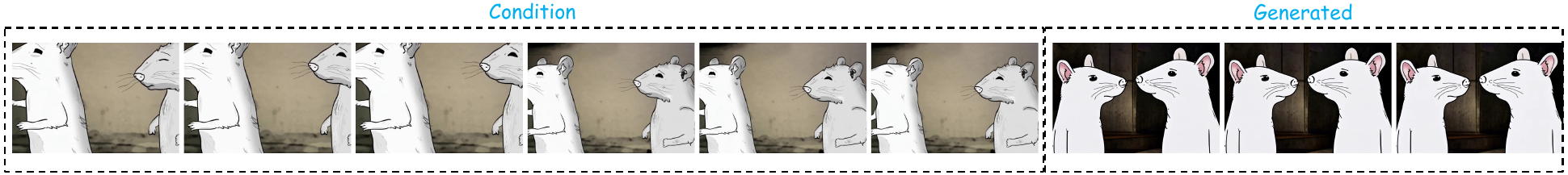}
        \caption{ 
        "Two animated mice, Actor1 and Actor2, stand in a dimly lit space with a rough, concrete-like backdrop. Actor1 is completely white, while Actor2 has light grey shading. Actor1 points towards something off-screen to the left.  Actor2 looks towards Actor1, its expression seemingly neutral.",
        "Two white rats face each other. The left rat blinks slowly, while the right one exhales visible puffs of air.",
        "Actor1 and Actor2, two anthropomorphic white rats, stand in a dimly lit area resembling a basement. They turn their heads towards each other."
    }
        \label{fig:subfig12}
    \end{subfigure}

    \begin{subfigure}[b]{\linewidth}
        \centering
        \includegraphics[width=\linewidth]{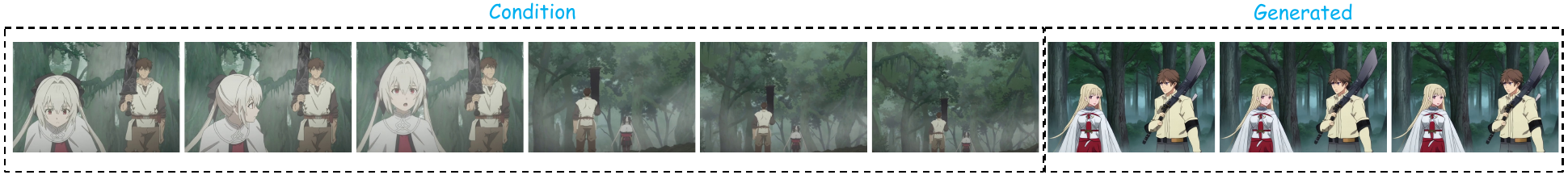}
        \caption{
        "A young woman with long, light hair and a white cloak stands in a misty forest with a young man carrying a large sword. She looks off to the side, seemingly curious or concerned.",
        "A man with brown hair, wearing a cream-colored, long-sleeved tunic and brown pants, and a pale-skinned woman with long white hair tied back with a dark ribbon, wearing a white and dark red outfit, stand at the edge of a misty forest with an uneven, foliage-littered floor.",
        "A woman with long, light hair, a white cape, and a red sash stands next to a man with brown hair and a cream shirt in a misty forest. He holds a large, dark sword over his shoulder. They both look to the left."
    }
        \label{fig:subfig13}
    \end{subfigure}
    \vspace{-7mm}
    \caption{Visualization results of auto-regressive scene extension.}
    \label{fig:ablation_show_maskclip}
\end{figure*}

\noindent\textbf{Auto-Regressive Scene Extension.}
\cref{fig:ablation_show_maskclip} and \cref{tab:ar} present the visualization and quantitative results of auto-regressive scene extension generated by Mask$^2$DiT, respectively, highlighting its effectiveness in extending videos based on a given set of visible scenes.
In the provided examples, the rightmost three columns display frames generated by Mask$^2$DiT.
These results highlight the model's ability to seamlessly extend scenes while maintaining temporal and semantic coherence.
Notably, the transitions between scenes are smooth, and the generated frames exhibit consistent visual quality, effectively capturing the intended narratives within each segment.
This showcases Mask$^2$DiT's capacity for generating longer, more complex videos with high fidelity.
As for the implementation details, we fix the timesteps $t$ corresponding to the visual token sequences of the conditional $n-1$ segments to 0, while the timesteps $t$ for the $n$-th segment to be generated remain consistent with those during standard training.
\vspace{-3mm}
\begin{table}[htbp]
\begin{center}
\resizebox{\linewidth}{!}{
\begin{tabular}{l c c c}
\hlineB{2}
Method  & Visual Con. (\%) & Semantic Con. (\%) & Sequence Con. (\%)\\
\hline\hline
Ours & 75.33 & 24.29 & 49.81 \\
\hlineB{2}
\end{tabular}
}
\end{center}
\vspace{-5mm}
\caption{Quantitative results of auto-regressive scene extension.}
\label{tab:ar}
\vspace{-5mm}
\end{table}

\noindent\textbf{More quantitative results.}
To further validate the effectiveness of Mask$^2$DiT, we additionally report evaluation results for both fixed-scene and auto-regressive scene extension settings on EvalCrafter~\cite{liu2024evalcrafter} and T2V-CompBench~\cite{sun2024t2v}, as shown in~\cref{tab:quantitative}.
Overall, our method demonstrates superior consistency and achieves competitive performance across multiple metrics on standard single-scene video generation benchmarks.
However, these benchmarks are not well-suited for our task.

\begin{table}[htbp]
\begin{center}
\resizebox{\linewidth}{!}{
\begin{tabular}{l c c c c c}
\hlineB{2}
Method  & Visual Con. (\%) & Semantic Con. (\%) & Sequence Con. (\%) & Aesthetic Quality & Imaging Quality \\
\hline\hline
Pre-training & 49.04 & \textbf{26.36} & 37.70 & - & -\\
SFT & 73.22 & 23.54 & 48.38 & 57.44 & 72.37 \\
Pre-training + SFT & \textbf{73.73} & 23.73 & \textbf{48.73} & \textbf{57.68} & \textbf{73.20} \\
\hlineB{2}
\end{tabular}
}
\end{center}
\vspace{-5mm}
\caption{Effect of pre-training dataset.}
\label{tab:sft_pretrained}
\vspace{-5mm}
\end{table}

\begin{table*}
\centering
\resizebox{\linewidth}{!}{
\begin{tabular}{c|c|ccc|cccc}
\hline
                             & Methods                        & VQA\_A         & VQA\_T         & flow score     & Action binding & Consistent Attr Binding & Dynamic Attr Binding & Object Interactions \\ \hline
scene extension              & Ours                           & 58.70          & 57.27          & 0.64           & 1.76           & 2.93                    & 0.0005               & 1.83                \\ \hline
\multirow{4}{*}{fixed scenes} & CogVideoX                      & {\ul 53.78}    & {\ul 63.76}    & {\ul 1.87}     & 2.16           & 2.84                    & {\ul 0.0076}         & 2.01                \\
                             & StoryDiffusion + CogVideoX-I2V & 28.71          & 33.96          & 1.08           & {\ul 2.17}     & {\ul 3.17}              & 0.0075               & \textbf{2.08}       \\
                             & TALC                           & 3.72           & 7.66           & \textbf{14.99} & 2.15           & 2.70                    & \textbf{0.0121}      & 2.01                \\
                             & Ours                           & \textbf{75.48} & \textbf{64.61} & 1.75           & \textbf{2.30}  & \textbf{3.28}           & 0.0065               & {\ul 2.02} \\
\hline
\end{tabular}
}
\vspace{-3.5mm}
\caption{Evaluation for auto-regressive scene extension and video generation with a fixed number of scenes on two T2V evaluation benchmarks, \textit{i.e.}, EvalCrafter~\cite{liu2024evalcrafter} and T2V-CompBench~\cite{sun2024t2v}.}
\label{tab:quantitative}
\vspace{-5mm}
\end{table*}

\noindent \textbf{Effect of Pre-training Dataset.}
To evaluate the impact of the pre-training dataset, we compare three configurations: using only the pre-training dataset, using only the SFT dataset, and a hybrid approach where the model is first pre-trained on the pre-training dataset and then fine-tuned with the SFT dataset, as shown in the Tab~\ref{tab:sft_pretrained}.
Since CogVideoX is trained on single-scene data, its extension to multi-scene generation often results in failures.
Pre-training allows the model to adapt to generating distinct scenes for different prompts.
However, because the pre-training dataset consists of randomly concatenated single-scene videos without inherent logical relationships, it exhibits lower consistency compared to results achieved with the SFT dataset.
Training exclusively on the SFT dataset improved Visual Consistency by 24.18\% over the pre-training dataset, underscoring the critical role of multi-scene consistency in data for our approach.
Finally, pre-training with the pre-training dataset followed by fine-tuning with the SFT dataset leads to a 0.35\% improvement in Sequence Consistency, highlighting the complementary benefits of pre-training, particularly in scenarios where SFT data is limited.
Furthormore, the pre-training phase contributes to enhanced visual quality in the generated videos, as evidenced by improvements in the aesthetic quality ($\textbf{57.68}$ vs. 57.44) and imaging quality ($\textbf{73.20}$ vs. 72.37) metrics reported by VBench~\cite{huang2024vbench}.
These findings suggest that incorporating additional high-quality, semantically coherent multi-scene data could further enhance the performance of our model.

\begin{table}
\begin{center}
\resizebox{\linewidth}{!}{
\begin{tabular}{l c c c c}
\hlineB{2}
Model  & Visual Con. (\%) & Semantic Con. (\%) & Sequence Con. (\%) & FVD ($\downarrow$) \\
\hline\hline
CogVideoX-2B & 55.01 & 22.64 & 38.82 & 835.35 \\
Ours & \textbf{70.95} & \textbf{23.94} & \textbf{47.45} & \textbf{720.01} \\
\midrule
CogVideoX-5B & 43.82 & 20.70 & 32.26 & 613.47 \\
Ours & \textbf{89.21} & \textbf{20.81} & \textbf{55.01} & \textbf{607.64} \\
\hlineB{2}
\end{tabular}
}
\end{center}
\vspace{-5mm}
\caption{Performance comparison with larger models. The 5B model improves multi-scene video generation by 7.56\% over the 2B model, highlighting the impact of stronger foundation models.}
\label{tab:5b}
\end{table}

\begin{table}[htbp]
    \begin{center}
    \resizebox{\linewidth}{!}{
        \begin{tabular}{l c c c c}
        \toprule
        Method  & Visual Con. (\%) & Semantic Con. (\%) & Sequence Con. (\%) & FVD ($\downarrow$) \\
        \hline\hline
        Ours(3s) & \textbf{87.62} & 22.38 & \textbf{55.00} & 818.96 \\
        Ours(6s) & 70.95 & \textbf{23.94} & 47.45 & \textbf{720.01} \\
        \bottomrule
        \end{tabular}
    }
    \end{center}
    \vspace{-6mm}
    \caption{Performance in shorter scene generation.}
    \label{tab:short-scene}
\end{table}

\begin{figure*}[h]
    \centering
    \includegraphics[width=\linewidth]{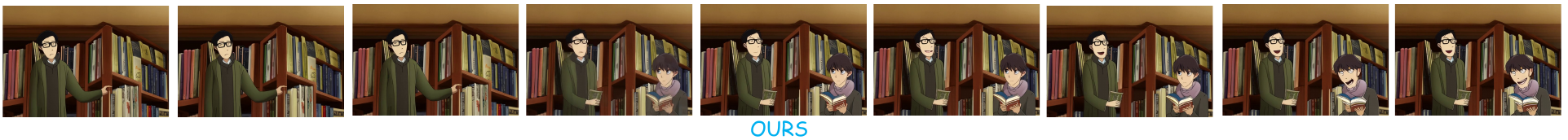}
    \vspace{-7mm}
    \caption{Visualization results of shorter scene generation.}
    \label{fig:short-scene}
\end{figure*}

\begin{figure*}
    \centering
    \includegraphics[width=0.96\linewidth]{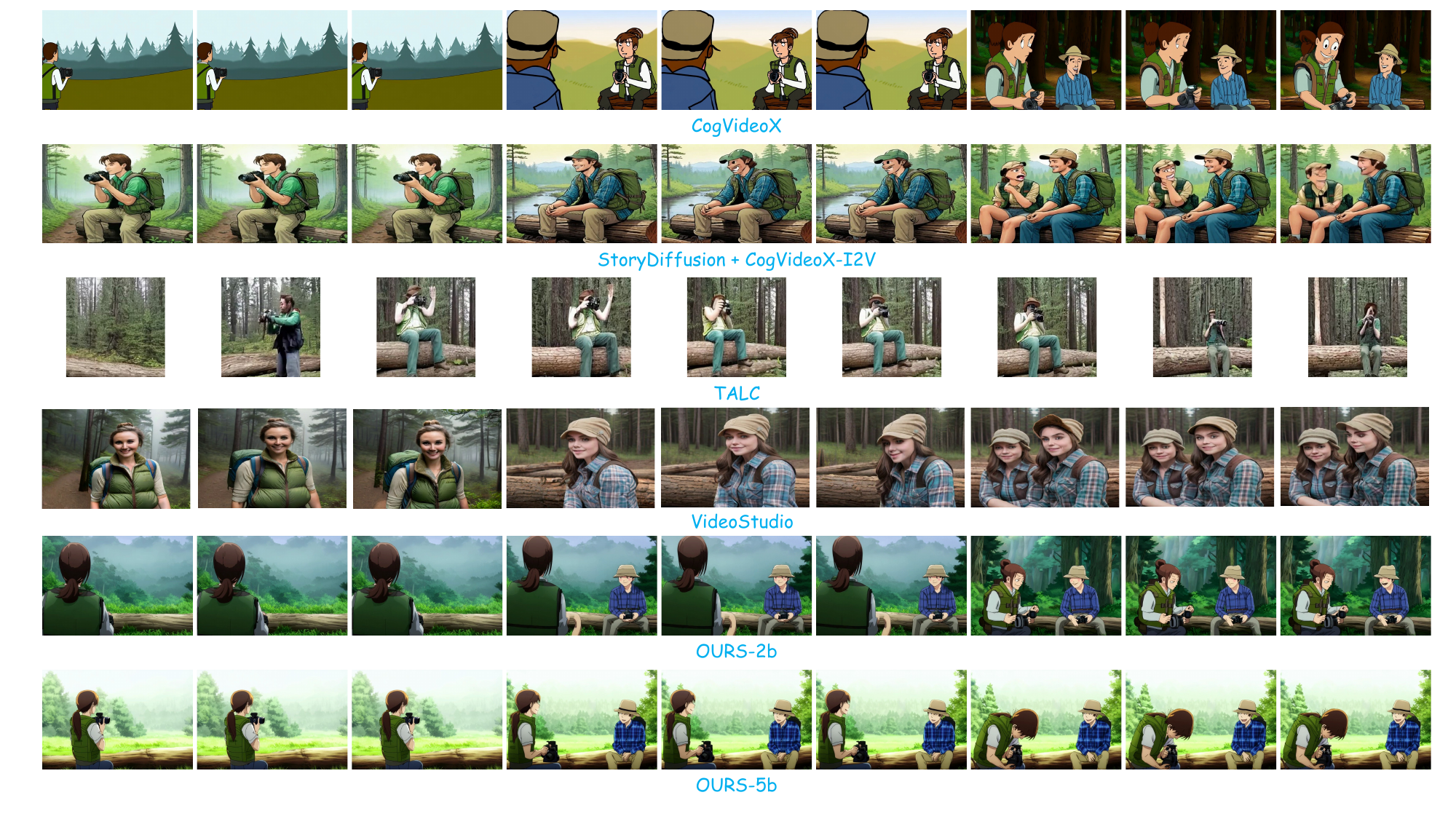}
    \vspace{-3mm}
    \caption{Visualization of Results from Different SOTA Methods. Video Captions: "Actor1 (brown hair tied back, wearing a green hiking vest, holding a camera) stands at the edge of the clearing, framing a shot of the misty forest.",
            "Actor2 (wearing a beige hat, blue flannel shirt, sitting on a log) watches Actor1 (brown hair tied back, wearing a green hiking vest, holding a camera) with a smile, appreciating the tranquility of the early morning.",
            "Actor1 (brown hair tied back, wearing a green hiking vest, holding a camera) lowers the camera and shares a quiet laugh with Actor2 (wearing a beige hat, blue flannel shirt, sitting on a log) as they both take in the serene beauty of the forest."}
    \label{fig:sota_5b1}
\end{figure*}

\begin{figure*}
    \centering
    \includegraphics[width=0.96\linewidth]{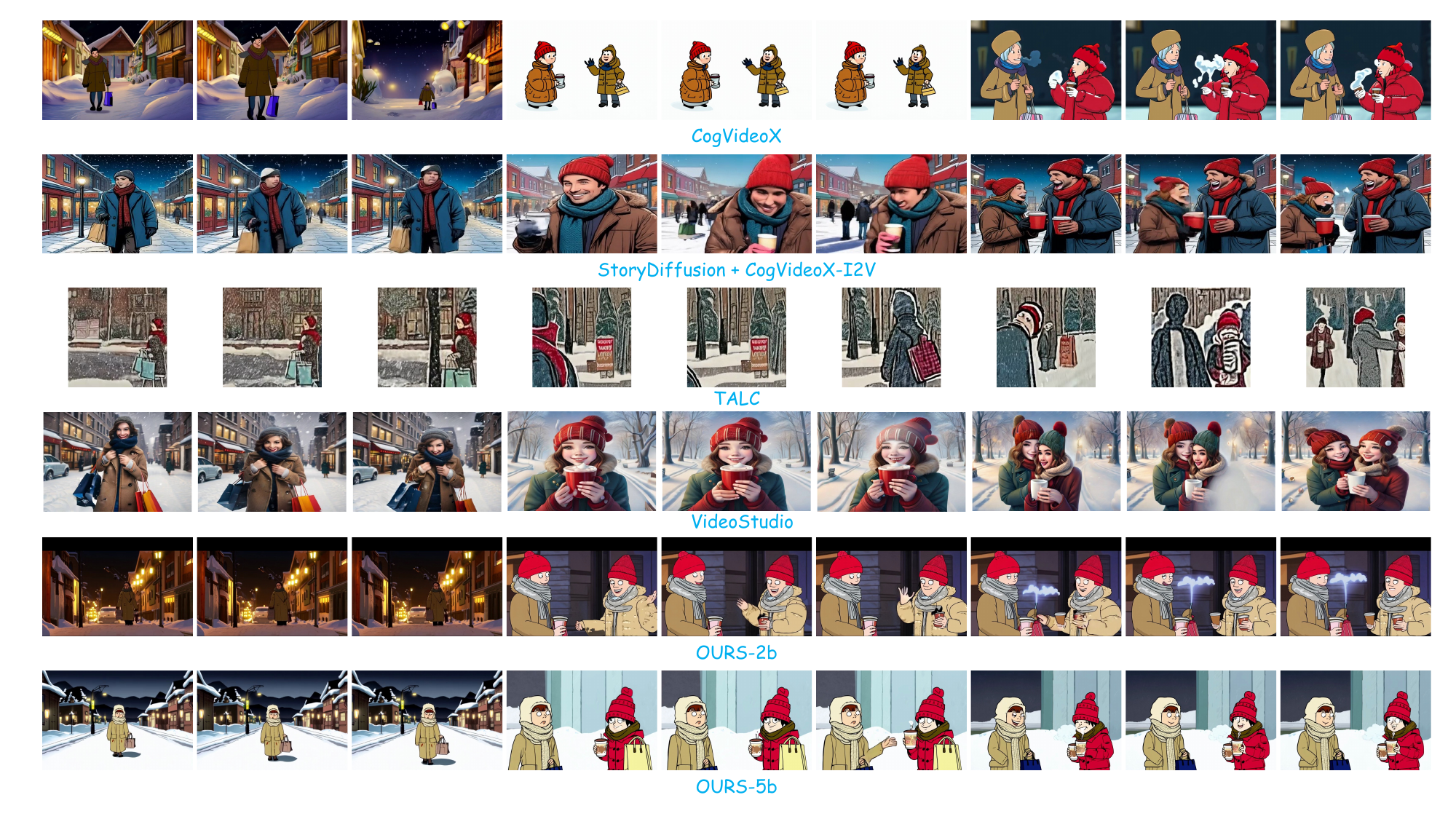}
    \vspace{-3mm}
    \caption{Visualization of Results from Different SOTA Methods. Video Captions: 
    "Actor1 (gray backpack, green jacket, reading a map) stands at the platform, studying the station layout.",
    "Actor2 (blonde ponytail, red scarf, holding a suitcase) notices Actor1 (gray backpack, green jacket, reading a map) and approaches to offer directions.",
    "Actor1 (gray backpack, green jacket, reading a map) thanks Actor2 (blonde ponytail, red scarf, holding a suitcase), both exchanging a friendly smile in the busy station."
    }
    \label{fig:sota_5b2}
\end{figure*}

\begin{figure*}
    \centering
    \includegraphics[width=0.96\linewidth]{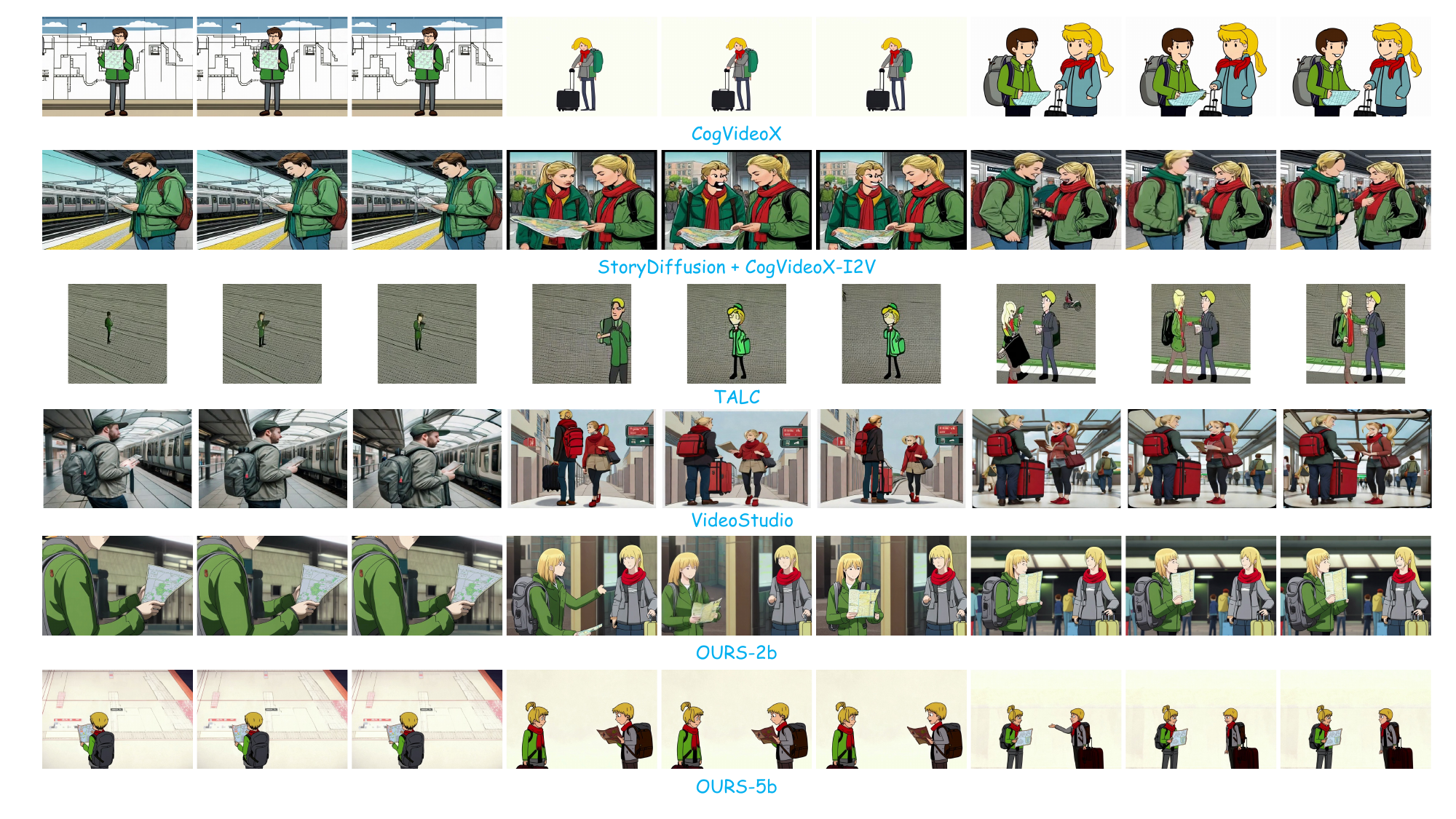}
    \vspace{-3mm}
    \caption{Visualization of Results from Different SOTA Methods. Video Captions: 
    "Actor1 (wavy hair, plaid shirt, playing a ukulele) strums a soft melody by the bonfire.",
    "Actor2 (braided hair, cozy blanket, singing along) joins in, creating a serene harmony under the stars.",
    "Actor1 (wavy hair, plaid shirt, playing a ukulele) laughs softly as Actor2 (braided hair, cozy blanket, singing along) harmonizes, both enjoying the peaceful night."
    }
    \label{fig:sota_5b3}
\end{figure*}

\begin{figure*}
    \centering
    \includegraphics[width=0.96\linewidth]{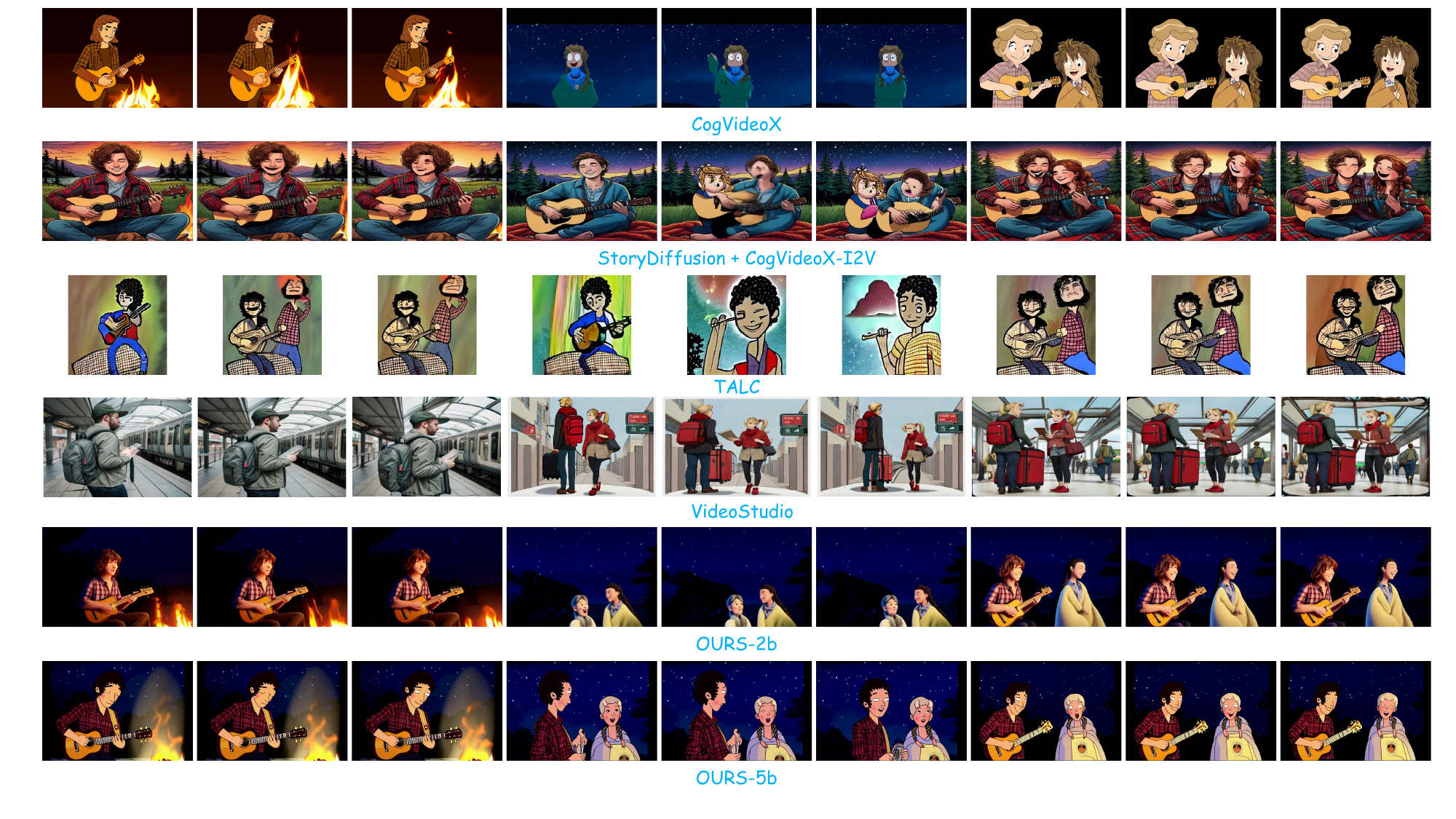}
    \vspace{-3mm}
    \caption{Visualization of Results from Different SOTA Methods. Video Captions: 
    "Actor1 (wavy hair, plaid shirt, playing a ukulele) strums a soft melody by the bonfire.",
    "Actor2 (braided hair, cozy blanket, singing along) joins in, creating a serene harmony under the stars.",
    "Actor1 (wavy hair, plaid shirt, playing a ukulele) laughs softly as Actor2 (braided hair, cozy blanket, singing along) harmonizes, both enjoying the peaceful night."
    }
    \label{fig:sota_5b4}
\end{figure*}

\begin{figure*}
    \centering
    \includegraphics[width=0.96\linewidth]
    {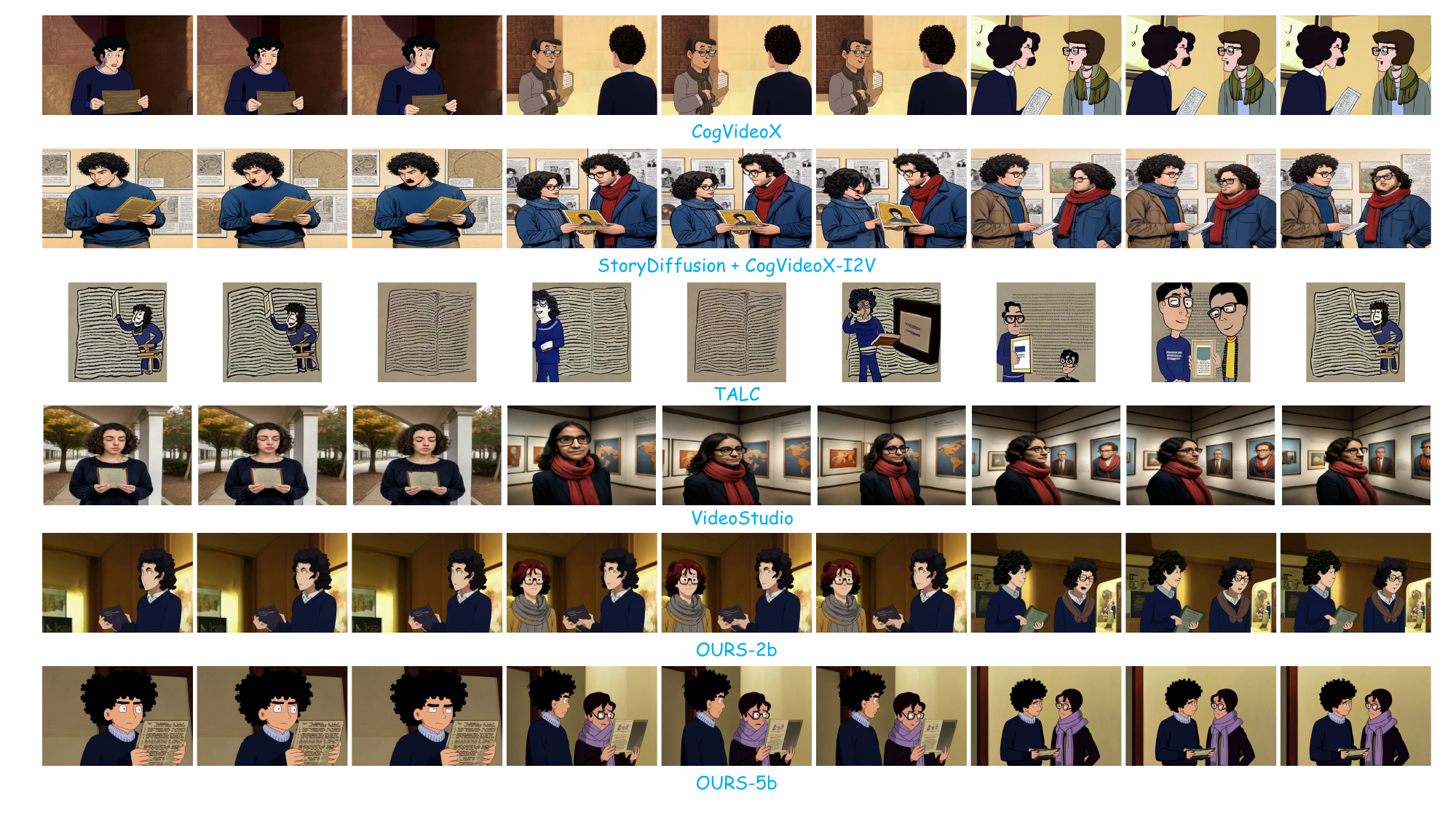}
    \vspace{-3mm}
    \caption{Visualization of Results from Different SOTA Methods. Video Captions: 
    "Actor1 (dark curly hair, navy sweater, reading a plaque) studies the information about an ancient artifact.",
    "Actor2 (wearing a scarf, glasses, admiring an exhibit) stands nearby, sharing an intrigued look with Actor1 (dark curly hair, navy sweater, reading a plaque).",
    "Actor1 (dark curly hair, navy sweater, reading a plaque) and Actor2 (wearing a scarf, glasses, admiring an exhibit) exchange thoughts about the exhibit, both captivated by the history around them."
    }
    \label{fig:sota_5b5}
\end{figure*}

\noindent \textbf{Performance Improvement with Larger Models.}
As shown in Tab~\ref{tab:5b}, we compare the performance of a 5B model and observe that improvements in the foundation model lead to significant gains in multi-scene video generation.
Specifically, the 5B model outperforms the 2B model with a 7.56\% improvement in Sequence Consistency.
To further illustrate the effectiveness of various foundation models, we present qualitative results in Fig.~\ref{fig:sota_5b1},~\ref{fig:sota_5b2},~\ref{fig:sota_5b3},~\ref{fig:sota_5b4},~\ref{fig:sota_5b5},~\ref{fig:sota_5b6},~\ref{fig:sota_5b7},~\ref{fig:sota_5b8}, with the original videos included in the supplementary compressed package.

\begin{table*}
  \centering
  \begin{tabular}{@{}lcccc@{}}
    \toprule
    Aspect & Visual Consistency$\uparrow$ & Semantic Consistency$\uparrow$ & Video Quality$\uparrow$ & Overall$\uparrow$ \\
    \midrule
    CogVideoX & 8.96 & 12.12 & 9.23 & 9.09 \\
    StoryDiffusion + CogVideoX-I2V & 29.85 & 28.79 & 29.23 & 28.79 \\
    TALC & 13.43 & 13.64 & 12.31 & 12.12 \\
    Ours & \textbf{47.76} & \textbf{45.45} & \textbf{49.23} & \textbf{50.00} \\
    \bottomrule
  \end{tabular}
  \caption{Results for the user study in percentages.}
  \label{tab:us}
\end{table*}

\noindent\textbf{User study.}
In addition to objective evaluations, we have also designed a user study to subjectively assess the practical performance of various methods.
Given 10 three-part prompts generated by ChatGPT, we adopt CogVideoX, StoryDiffusion equipped with CogVideoX-I2V, TALC, and Mask$^2$DiT to generate corresponding multi-scene videos separately.
Each method is executed once with the same random seed, ensuring a fair comparison by eliminating randomness in generation.
We ask 10 users from distinct backgrounds to evaluate the generated results across 4 dimensions based on the following question: ``For a given three-scene text prompt, the following options present results from different multi-scene long video generation methods.
Please evaluate the results in terms of text-video alignment (faithfulness to the given text prompt), video quality, and visual consistency across scenes.
Based on these three criteria, please select the best result for each aspect and, finally, choose the overall best result considering all three aspects."
Since abstention is allowed, we ultimately receive 266 valid votes. 
The final results are displayed in~\cref{tab:us}.
Mask$^2$DiT outperforms all state-of-the-art methods in all three evaluation aspects and overall preference by a big margin, demonstrating the broad application prospects of our method.

\noindent\textbf{Ablation studies on probability $p$.}
We add ablation studies on probability $p$, as shown in Tab~\cref{tab:ab_p}.

\begin{table*}
\centering
\resizebox{\linewidth}{!}{
\begin{tabular}{c|ccc|ccc|c}
\hline
     & \multicolumn{3}{c|}{Video generation with a fixed number of scenes}                  & \multicolumn{3}{c|}{Auto-regressive scene extension}                  & Overall        \\ \hline
$p$ & Visual Con.    & Semantic Con.  & Sequence Con.  & Visual Con.    & Semantic Con.  & Sequence Con.  & Sequence Con.  \\ \hline
0.1  & 72.37          & 23.36          & 47.86          & 72.63          & 23.38          & 48.01          & 47.94          \\
0.3  & \textbf{73.51} & 23.61          & \textbf{48.56} & 73.89          & 23.87          & 48.88          & 48.72          \\
0.5  & 73.21          & 23.82          & 48.52          & 75.33          & \textbf{23.91} & 49.62          & \textbf{49.07} \\
0.7  & 71.63          & \textbf{23.92} & 47.78          & \textbf{77.57} & 22.81          & \textbf{50.19} & 48.98          \\
0.9  & 67.96          & 23.54          & 45.75          & 75.73          & 23.50          & 49.61          & 47.68          \\ \hline
\end{tabular}
}
\caption{Ablation studies on the probability $p$.}
\label{tab:ab_p}
\end{table*}

\noindent\textbf{Ablation studies on generating shorter sequences.}
We directly apply our model trained on 6-second scenes to generate 3-second scenes, obtaining promising results, as shown in~\cref{tab:short-scene} and~\cref{fig:short-scene}.
The only limitation is a slight degradation in semantic consistency.

\noindent\textbf{Limitations.}
The limitations are twofold. First, due to the constraints of the training data, our model is limited to generating animated videos. Second, the motion dynamics and durations for each scene require further investigation.

\vspace{-10mm}

\begin{figure*}
    \centering
    \includegraphics[width=0.96\linewidth]{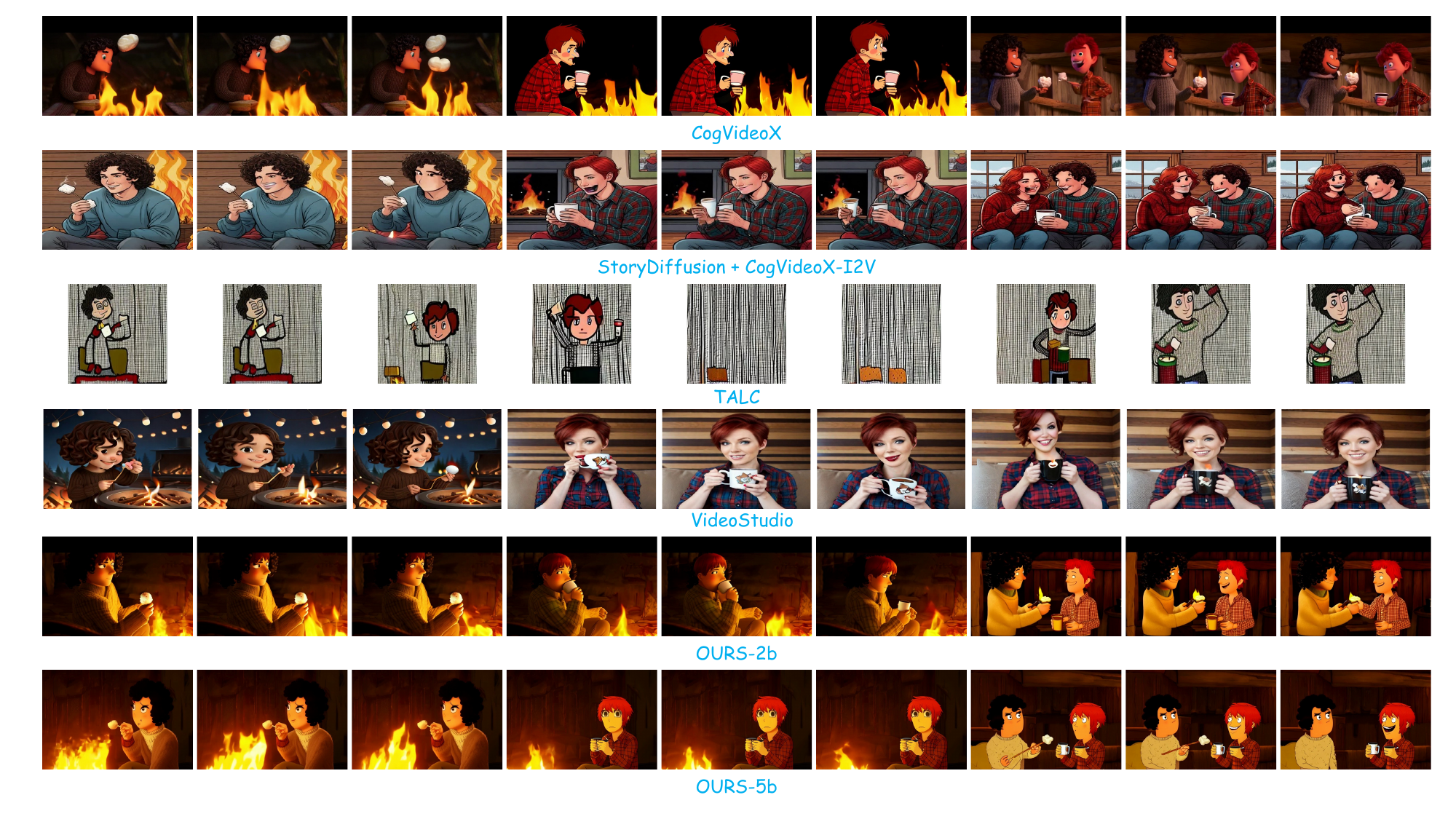}
    \vspace{-3mm}
    \caption{Visualization of Results from Different SOTA Methods. Video Captions: 
    "Actor1 (dark curly hair, cozy sweater, roasting marshmallows) holds a marshmallow over the flames, watching it toast.",
    "Actor2 (short red hair, plaid shirt, holding a mug of hot cocoa) sits nearby, savoring the warmth of the fire.",
    "Actor1 (dark curly hair, cozy sweater, roasting marshmallows) offers a marshmallow to Actor2 (short red hair, plaid shirt, holding a mug of hot cocoa), both sharing laughter in the cozy cabin."
    }
    \label{fig:sota_5b6}
\end{figure*}

\begin{figure*}
    \centering
    \includegraphics[width=0.96\linewidth]{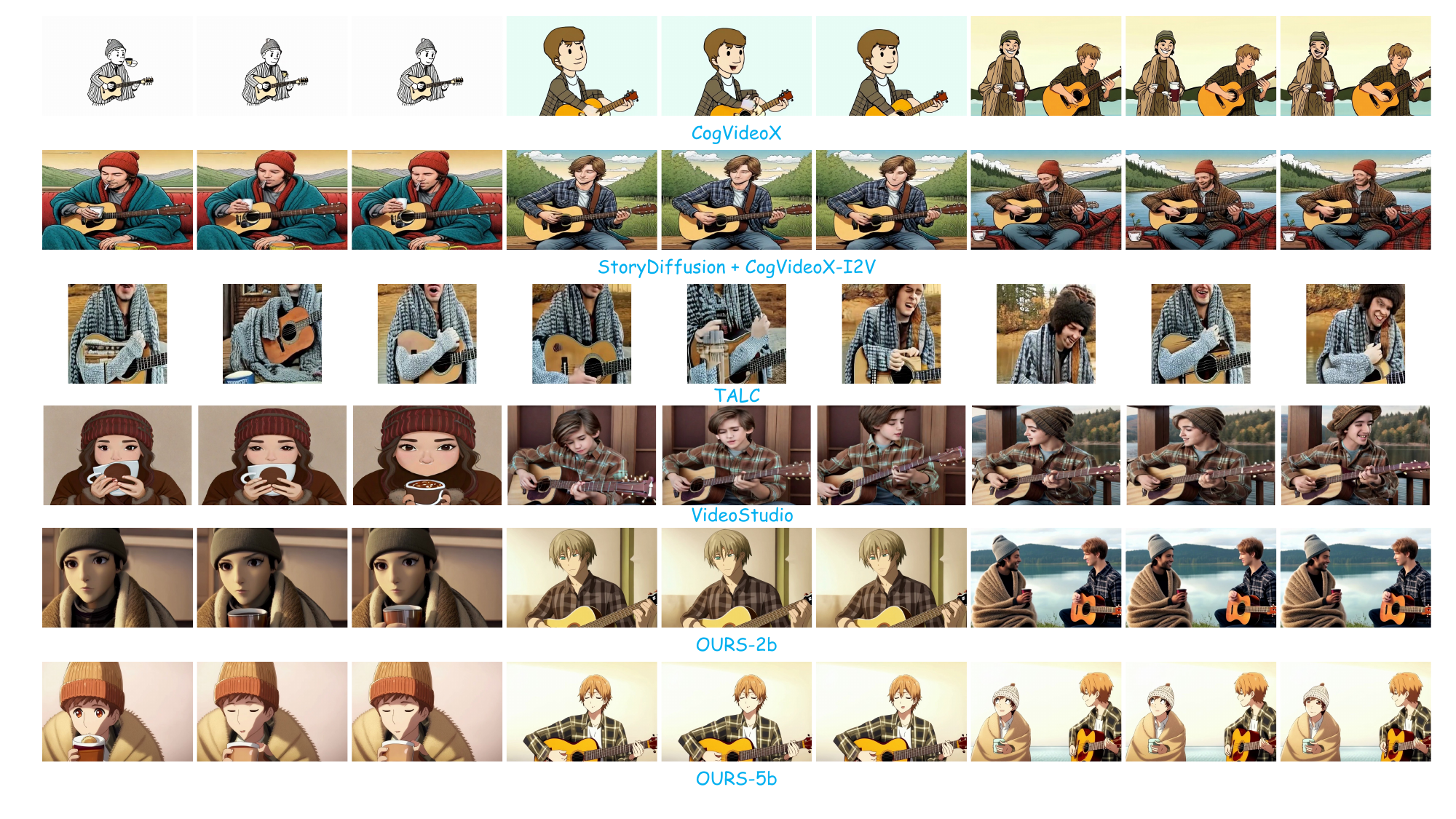}
    \vspace{-3mm}
    \caption{Visualization of Results from Different SOTA Methods. Video Captions: 
    "Actor1 (wearing a beanie, wrapped in a wool blanket, holding a cup of tea) sips their drink, listening to the soft sounds of the guitar.",
    "Actor2 (light brown hair, in a flannel shirt, strumming a guitar) plays a gentle tune, filling the peaceful air.",
    "Actor1 (wearing a beanie, wrapped in a wool blanket, holding a cup of tea) smiles and relaxes, enjoying the moment with Actor2 (light brown hair, in a flannel shirt, strumming a guitar) by the lake."
    }
    \label{fig:sota_5b7}
\end{figure*}

\begin{figure*}[htbp]
    \centering
    \includegraphics[width=0.96\linewidth]{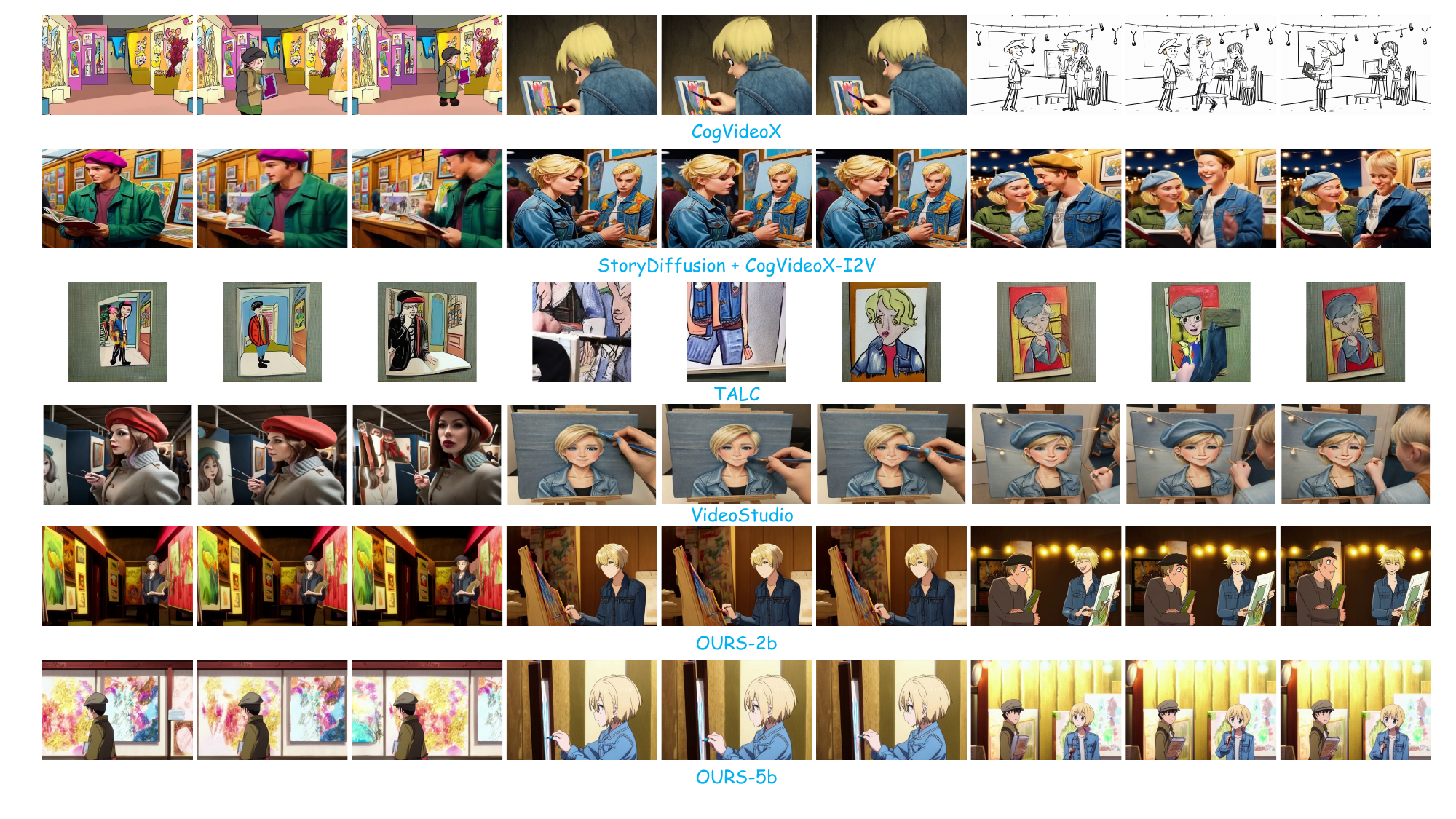}
    \vspace{-3mm}
    \caption{Visualization of Results from Different SOTA Methods. Video Captions: 
    "Actor1 (wearing a beret, carrying a sketchbook, admiring artwork) walks through the booths, captivated by the vibrant colors.",
    "Actor2 (short blonde hair, denim jacket, painting on a small canvas) concentrates, adding finishing touches to their piece.",
    "Actor1 (wearing a beret, carrying a sketchbook, admiring artwork) stops to compliment Actor2 (short blonde hair, denim jacket, painting on a small canvas), both exchanging smiles under the string lights."
    }
    \label{fig:sota_5b8}
\end{figure*}

\end{document}